\documentclass[final,3p,times]{elsarticle}

\usepackage{amssymb}

\usepackage{lineno}

\usepackage[utf8]{inputenc} 
\usepackage[T1]{fontenc}    
\usepackage[hidelinks]{hyperref}       
\usepackage{url}            
\usepackage{booktabs}       
\usepackage{amsfonts}       
\usepackage{nicefrac}       
\usepackage{microtype}      
\usepackage{graphicx}
\usepackage{natbib}
\setcitestyle{square,numbers,sort&compress}
\biboptions{numbers,sort&compress}
\usepackage{doi}
\usepackage{color,soul}
\usepackage{multirow}
\usepackage{longtable}
\usepackage{tikz}
\usetikzlibrary{positioning, shapes.geometric, backgrounds}

\journal{Elsevier}

\begin{document}

\begin{frontmatter}



    \title{A Systematic Literature Review of Computer Vision Applications in Robotized Wire Harness Assembly}


    \author[a]{Hao Wang\corref{cor1}}

    \author[a]{Omkar Salunkhe}

    \author[b]{Walter Quadrini}

    \author[c]{Dan L{\"a}mkull}

    \author[d]{Fredrik Ore}

    \author[a]{M{\'e}lanie Despeisse}

    \author[b]{Luca Fumagalli}

    \author[a]{Johan Stahre}

    \author[a]{Bj{\"o}rn Johansson}

    \affiliation[a]{
    organization={Department of Industrial and Materials Science, Chalmers University of Technology},
    addressline={H{\"o}rsalsv{\"a}gen 7A},
    city={Gothenburg},
    postcode={SE-412 96},
    country={Sweden}
    }

    \affiliation[b]{
        organization={Department of Management, Economics and Industrial Engineering, Politecnico di Milano},
        addressline={Piazza Leonardo da Vinci 32},
        city={Milan},
        postcode={20133},
        country={Italy}
    }

    \affiliation[c]{
        organization={Department of Innovation and Strategy, Volvo Car Corporation},
        city={Gothenburg},
        postcode={SE-418 78},
        country={Sweden}
    }

    \affiliation[d]{
    organization={Department of Global Industrial Development, Scania CV AB},
    city={S{\"o}dert{\"a}lje},
    postcode={SE-151 87},
    country={Sweden}
    }

    \cortext[cor1]{Corresponding author. Tel.: +46-(0)31-772-1202. \textit{E-mail address}: haowang@chalmers.se}

    \begin{abstract}


        In the current automotive industry, human operators perform wire harness assembly manually, which causes significant quality, productivity, safety, and ergonomic problems.
        Robotic assembly is a critical facilitator in addressing these problems.
        However, it remains challenging to implement for robotizing the assembly of wire harnesses.
        Wire harness assembly is a specific scenario of deformable linear object manipulation.
        Robotizing this assembly task demands robots to flexibly adapt their actions to the dynamically changing industrial environment based on robotic perception results.
        Existing research suggested the significance of robotic visual perception in the robotic assembly of wire harnesses.
        Implementing computer vision techniques is fundamental to enabling robots' visual perception capabilities.
        Nonetheless, the industry has yet to introduce vision-based solutions to robotize wire harness assembly fully or partially.
        Through a systematic literature review, this article identifies $15$ scientific publications in vision-based robotized wire harness assembly.
        The results show various computer vision applications regarding wire harness components and assembly operations studied in previous research.
        Nevertheless, this article recognizes two significant challenges for computer vision applications in robotized wire harness assembly: 1) fulfilling production requirements on robustness and practicality and 2) exploiting the intrinsic physical features of wire harnesses for visual recognition.
        This article also advocated five prospective research directions toward more efficient and practical vision-based robotized wire harness assembly: 1) developing learning-based vision systems to exploit intrinsic features and multi-modality data of wire harnesses; 2) adapting vision systems proposed for robotizing assembly operations in manufacturing wire harnesses; 3) assessing the practicality, robustness, reliability, and sustainability of vision systems; 4) inquiring vision-based human-robot collaboration; and 5) exploring new product designs for facilitating visual recognition.

    \end{abstract}






\end{frontmatter}

\section{Introduction}
\label{sec:intro}

The assembly of wire harnesses in the final assembly station in the automotive industry has been performed entirely manually by skilled human operators over time.
Industry has identified this manual assembly as one of the bottlenecks constraining the promotion of automobile production remarkably~\cite{zagar2023copy}.
The manual assembly operations also jeopardize assembly quality, safety, and ergonomics in the contemporary automotive industry~\cite{hermansson2013automatic,salunkhe2023review}.
Significantly, the rapidly growing demand for electric vehicles (EV) exacerbates the production problems amid the massive industrial and social transformation toward autonomous driving, electrification, and green transportation~\cite{ec2021european,eea2022decarbonising,fankhauser2022meaning}.
Implementing automation, particularly robotic assembly, is one of the prominent approaches to address these problems in actual productions~\cite{riley1996assembly,boothroyd2005assembly,hu2011assembly}.

The robotic assembly has been implemented to facilitate automation in various industries, accomplished by robots solely or human-robot collaboration (HRC)~\cite{chen2015robotic,leng2022industry}.
Robotic assembly is favored to fulfill the increasingly demanding wire harness assembly in the automotive industry~\cite{salunkhe2023review}.
The robotic assembly enables more rigorous, safer, and more ergonomic-friendly manufacturing than manual operations due to its better replicability, transparency, and comprehensibility~\cite{mason2018robotic}.
Various preliminary automation solutions for wire harness assembly were proposed under simplified industrial configurations decades ago~\cite{saadat2002industrial}.
The past few years have also witnessed studies proposing various robotized solutions for wire harness assembly in different industrial sectors, e.g., electronics, automobile, and aviation~\cite{salunkhe2023review}.
However, a low level and limited scale of automation on wire harness assembly in actual automobile production is observed, and robotic assembly of wire harnesses remains arduous to achieve in actual circumstances~\cite{makris2023automated}.

The conventional robotic automation solutions struggled with the task of wire harness assembly.
The challenges stem from the high complexity of the assembly tasks, regarding, e.g., the numerous variants of wire harnesses to be assembled, the complex object deformation and dynamics~\cite{rambow2012autonomous}, and the limited process time in actual production~\cite{shi2012mobile}.
Before assembling wire harnesses robotically, robots require more advanced perception capability to recognize wire harness components' positions and orientations and track wire harnesses' movement and deformation.
Vision is instrumental for object localization, classification, and tracking~\cite{billard2019trends}.

Research in computer vision and machine vision has demonstrated to the automotive industry the potential to facilitate robotized wire harness assembly.
Visual inputs embed vast information of the outer environment~\cite{martinez2019vision}.
Research on robotic manipulation reveals the significance of vision systems for robotic perception~\cite{shahria2022comprehensive}.
Moreover, numerous studies have investigated diverse computer vision applications in different manufacturing scenarios~\cite{zhou2023computer}.
There has also been discussion on facilitating better robotic visual perception and manipulation of wire harnesses with computer vision techniques in previous research~\cite{wang2023overview}.
Nonetheless, no such practical automated solutions for the final assembly of wire harnesses have yet to be witnessed in the actual production~\cite{makris2023automated}.
The overall robotic assembly of automotive wire harnesses enabled by vision systems remains unsolved.
The full potential of vision-based solutions in the robotized wire harness assembly remains unrevealed.
Hence, it is necessary to recognize existing challenges and potential remedies for developing computer vision-enabled robotized wire harness assembly.

This study aimed to summarize the state-of-the-art research on applying computer vision techniques to facilitate the robotic assembly of wire harnesses, identify the challenges for vision systems, and propose future research directions for developing a more practical vision-based robotic assembly of wire harnesses.
Regarding these aims, this study intended to answer the following research questions (RQ) through a systematic literature review:

\begin{enumerate}
    \item What computer vision-based solutions have been proposed for robotized wire harness assembly?
    \item What are the challenges for computer vision applications in robotized wire harness assembly?
    \item What are the required future research activities and fields for developing more efficient and practical computer vision-based robotized wire harness assembly?
\end{enumerate}

For clarification, the ``wire harness assembly'' discussed in this article is defined as the final installation of wire harnesses onto other products, e.g., installing wire harnesses onto electric vehicles in the final assembly of automobiles, instead of manufacturing wire harnesses, which has been reviewed in other studies~\cite{trommnau2019overview,nguyen2021manufacturing,navas2022wire}.

This article is organized as follows:
Section~\ref{sec:intro} introduces the systematic literature review's background and research questions.
Section~\ref{sec:bgd} describes the need for robotic assembly, the related research in robotic manipulation of deformable linear objects (DLO), the current assembly operations of automotive wire harnesses, and the challenges of automating the assembly.
Section~\ref{sec:method} describes the methodology implemented for the systematic literature review.
Section~\ref{sec:result} summarizes the latest advances in computer vision techniques implemented in robotized wire harness assembly, followed by discussions on current challenges and opportunities for future studies in Section~\ref{sec:disc}.
Section~\ref{sec:con} concludes this article with an outlook of future trends and research.

\section{Robotized automotive wire harnesses assembly}
\label{sec:bgd}

\subsection{Why is robotic assembly needed for wire harness assembly?}
\label{subsec:why-needed}

\begin{figure}[htb]
    \centering
    \includegraphics[width=\linewidth]{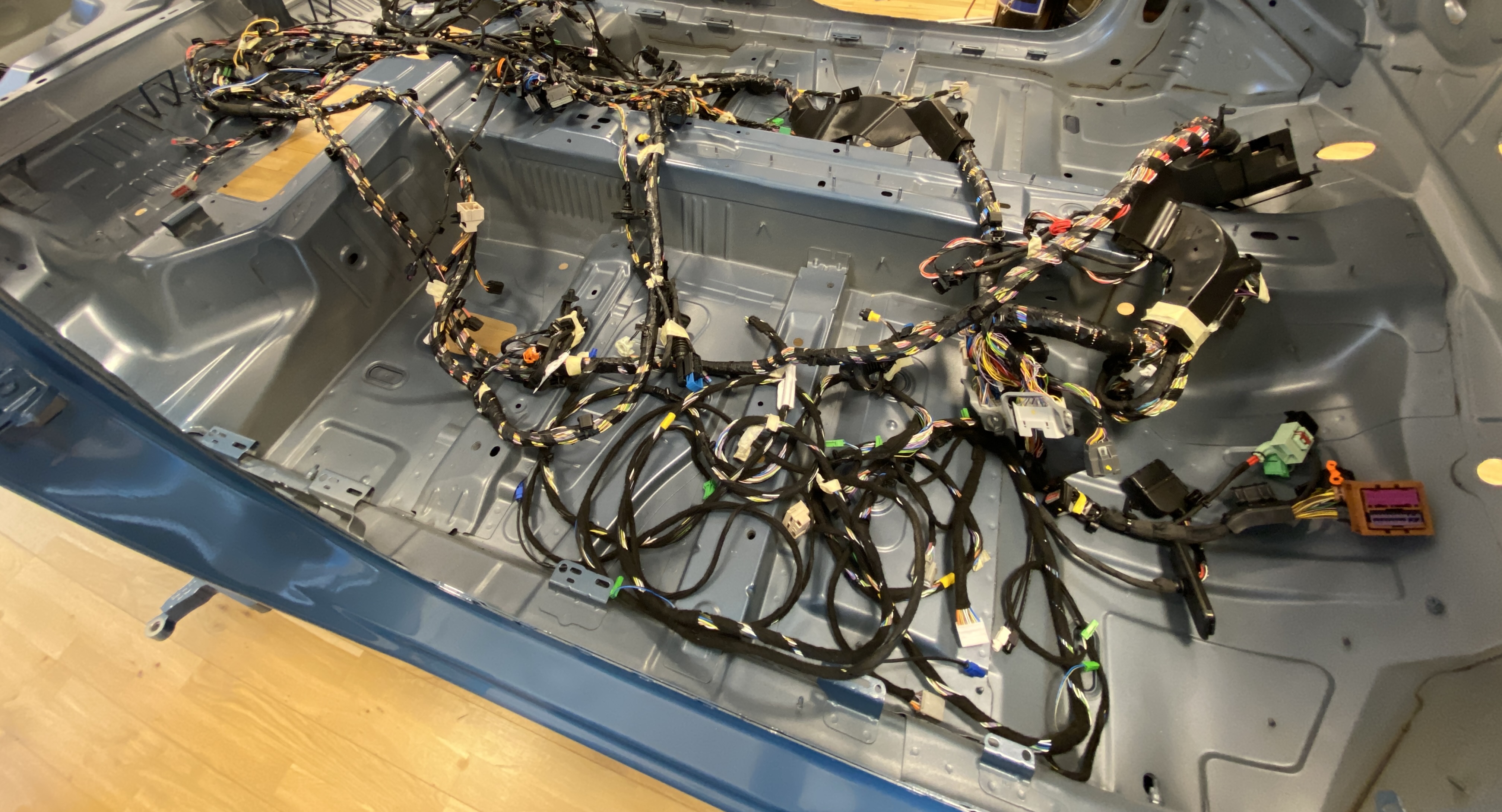}
    \caption{
    An example of the floor harnesses to be installed into passenger cabins of automobiles.
    (Courtesy of Volvo Car Corporation)
    \textbf{Color should be used for this figure in print and size should be calibrated for the camera-ready version (This sentence will be removed for the final submission).}
    }
    \label{fig:wire_example}
\end{figure}

\begin{figure}[htb]
    \centering
    \includegraphics[width=\linewidth]{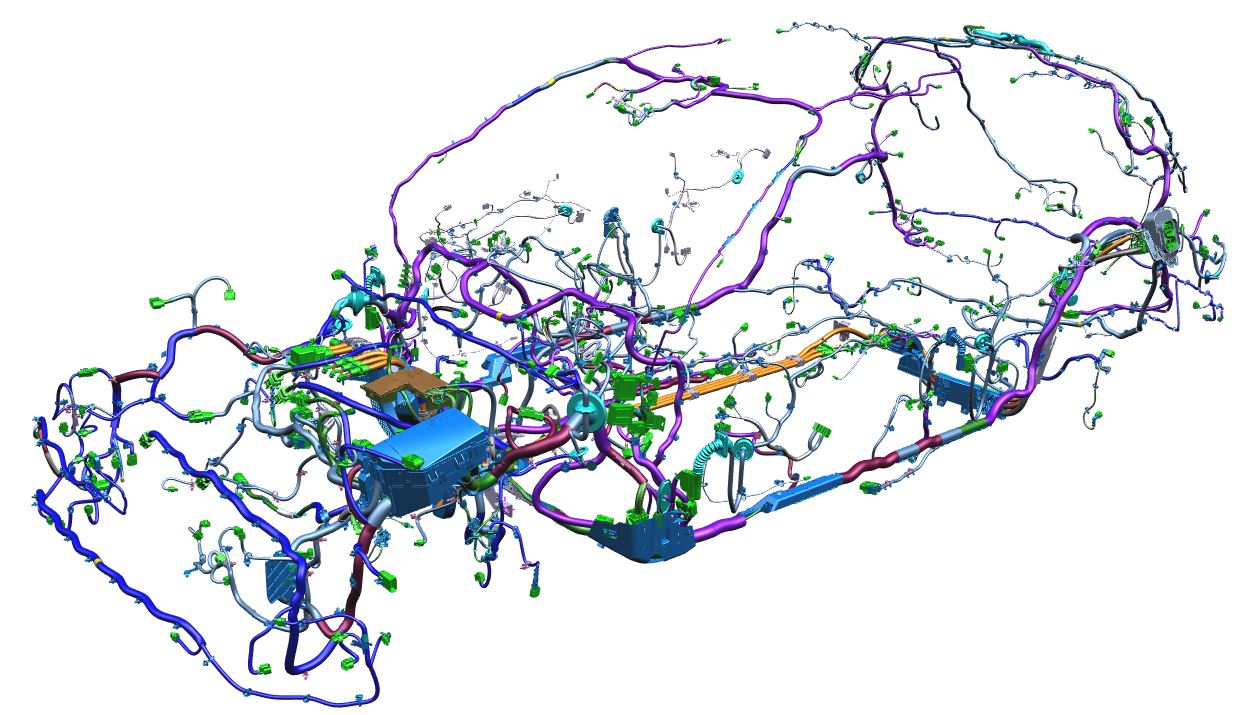}
    \caption{
    The electrical infrastructure of a Volvo XC40 Recharge, consisting of front, central, cabin, and rear cable systems.
    (Courtesy of Volvo Car Corporation)
    \textbf{Color should be used for this figure in print and size should be calibrated for the camera-ready version (This sentence will be removed for the final submission).}
    }
    \label{fig:elec_infra}
\end{figure}

It is essential to guarantee a high-quality assembly of wire harnesses in automobiles.
A wire harness is a bundle of routed cables and wires with a tree-like structure, consisting of numerous components, e.g., wires, terminals, connectors, clamps, and wrapping materials~\cite{aguirre1994economic}.
Figure~\ref{fig:wire_example} presents an example of the floor harnesses to be assembled into passenger cabins.
Wire harnesses are distributed extensively in modern automobiles, as the electrical infrastructure of a \textit{Volvo XC40 Recharge} illustrated in Figure~\ref{fig:elec_infra}, 
They are fundamental elements within automotive electronic systems responsible for quality-essential functions (e.g., engine control unit and energy transmission system) and safety-critical processes (e.g., maneuvering, driver assistance, and safety system)~\cite{nguyen2021manufacturing}.
Through correctly connected wire harnesses, signals and the current of electricity are transmitted among different electrical components scattered throughout electrical equipment to enable the overall system to function properly~\cite{aguirre1994economic,tilindis2014effect}.
However, the fully manual assembly of wire harnesses into vehicles causes problems regarding assembly quality due to the inevitable inconsistency of manual operation quality~\cite{hermansson2013automatic,wnuk2021tracking}.

\begin{figure}[htbp]
    \centering
    \includegraphics[width=\linewidth]{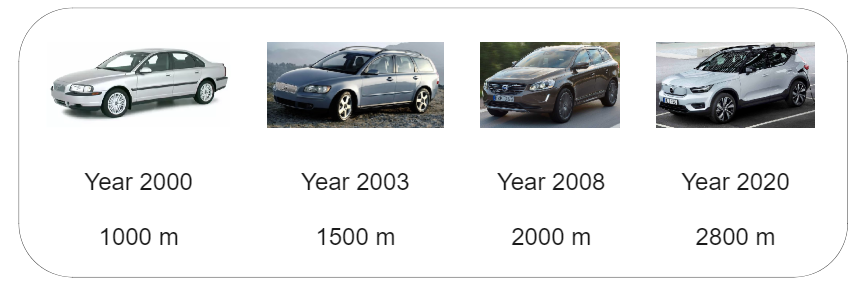}
    \caption{
    The remarkably increasing length of wires in passenger cars over time.
    (Courtesy of Volvo Car Corporation)
    \textbf{Color should be used for this figure in print and size should be calibrated for the camera-ready version (This sentence will be removed for the final submission).}
    }
    \label{fig:increasing_length}
\end{figure}

The efficiency of wire harness assembly is also instrumental in production, considering the increasing number of wire harnesses installed in modern vehicles within limited cycle time.
The usage of wire harnesses in modern vehicles has been enlarging remarkably over time~\cite{zagar2023copy}.
Figure~\ref{fig:increasing_length} illustrates an example of increasing usage of wire harnesses from the sector of passenger vehicles.
Industry also anticipates the continuous increment of wire harnesses installed in future automobiles~\cite{strategic2023automotive}, especially considering the growing number of electronic devices installed for various functions and the shift toward autonomous driving, electrification, and more sustainable mobility in the automotive industry~\cite{eea2022decarbonising,fankhauser2022meaning}.
Meanwhile, the automotive industry persists in a continuous demand to promote competitiveness and acquire market share, which lays a consistent requirement on promoting productivity.
The manual assembly, though, constrains the promotion of the overall productivity~\cite{riley1996assembly}.

Moreover, it is crucial to ensure human operators' safety and improve the wire harness assembly's ergonomics.
A large proportion of the manual operations in current wire harness assembly are skill-demanding and not ergonomic for human operators, e.g., heavy lifting (e.g., approximately $40$ kg for some automotive wire harnesses), high-pressure pressing, and far-reaching operation~\cite{salunkhe2023review}.
These manual operations cause musculoskeletal disorders (MSD) and occupational safety and health (OSH) issues in the workforce~\cite{trommnau2019overview}.
Adapting exoskeleton or other powered mechanics can enhance the physical strength of human operators~\cite{sylla2014ergonomic,romero2016towards}.
However, the assembly problems due to manual operations cannot be addressed effectively.
In addition, high-voltage wire harnesses are installed in automobiles, especially in electric vehicles.
These components demand meticulous material handling regarding safety, assembly quality, and reliability~\cite{fischer2015worker,olbrich2022manufacturing}.
Therefore, it is desired to ensure assembly quality and safety, promote productivity, improve ergonomics, and optimize resource utilization.
Implementing automation and robotic assembly is a prominent approach~\cite{riley1996assembly,boothroyd2005assembly,hu2011assembly}.

\subsection{Robotic manipulation of deformable linear objects}
\label{subsec:DLO}

Wire harnesses can be theoretically generalized as deformable linear objects (DLO)~\cite{galassi2021robotic,lv2022dynamic,makris2023automated}.
Wire harness assembly, thus, can be regarded as a specific task of DLO manipulation~\cite{saadat2002industrial}.

The robotics community has extensively devoted to addressing robotic manipulation of deformable objects~\cite{sanchez2018robotic,yin2021modeling,zhu2022challenges,gu2023survey}, also called non-rigid objects~\cite{nadon2018multi,makris2023automated} or flexible materials~\cite{saadat2002industrial}.
Deformable object manipulation widely exists in various application scenarios~\cite{saadat2002industrial}, e.g., manufacturing industry~\cite{li2018vision}, food industry~\cite{gemici2014learning}, medical surgery~\cite{cao2020sewing}, and daily activity~\cite{zhang2022learning}.

\textit{Deformable objects} with ``one dimension significantly larger than the other two'' are defined as DLOs~\cite{sanchez2018robotic}, e.g., ropes and cables.
Deformable linear objects are also called deformable one-dimensional objects (DOO)~\cite{javdani2011modeling,keipour2022deformable}.
DLO manipulation has also been a significant concern in industry over time~\cite{saadat2002industrial,sanchez2018robotic,makris2023automated}, e.g., the wire insertion in the electrical industry~\cite{chen1991deformation,wang2015online} and the assembly of cables in the automotive industry~\cite{byun1996determining,navas2022wire}.

Robotic manipulation of DLOs involves various robotic tasks, e.g., modeling, perception, and manipulation~\cite{sanchez2018robotic,nadon2018multi,yin2021modeling}.
The robotics community has been continuously investigating different models of deformable objects and the integration of various sensors and artificial intelligence (AI) into robots to equip robots with fast, accurate, and multi-modal perception abilities~\cite{lee2020making} and adaptive modeling and control abilities~\cite{yin2021modeling}.
As a prerequisite for accomplishing complex manipulation tasks, robotic perception is critical and required to perceive DLOs' shape, topology, deformation, and other physical properties before and during the robotic manipulation~\cite{sanchez2018robotic,lee2020making,zhu2022challenges}.
Robotic DLO manipulation essentially involves visual and tactile perception through visual, sound, force, tactile, and range sensing, independently or jointly~\cite{nadon2018multi,yin2021modeling,zhu2022challenges,makris2023automated}.
Mainly in existing research, visual perception was employed to obtain global information about DLOs' shapes on a large scale, especially when objects exhibit deformation on a large scale~\cite{nair2017combining,yan2020self}.
The tactile perception was often involved in acquiring shape and contact information on the local level~\cite{zhu2022challenges}.



Regardless of the tremendous advancement of robotics, DLO manipulation remains challenging in the robot-centered flexible automation~\cite{zhou2020practical,andronas2022perception}.
In particular, challenges in DLO manipulation remain in object detection, object modeling, deformation state estimation, and robotic operation~\cite{hermansson2013automatic,sanchez2018robotic,yin2021modeling,zhu2022challenges,andronas2022perception}.
Additionally, strategies designed for manipulating regular rigid objects cannot be adapted for DLO manipulation directly due to the high degrees of freedom and deformability of DLOs~\cite{sanchez2018robotic,zhou2020practical,wnuk2021tracking}.
These challenges call for further research and development in both theory and application aspects.

\subsection{General automation challenges in manufacturing}

The third industrial revolution initiated the broad adoption of automation in various sectors of industry~\cite{leng2022industry}.
However, many challenges still hamper the scale-up of automation applications in manufacturing.

Safety is fundamental in manufacturing, especially when robots are deployed in the production line~\cite{iso10218part1,iso10218part2,iso15066}.
Typically, different devices, e.g., steel fences and laser curtains, are installed around the working area of robots to guarantee safety by keeping human operators at a safe distance from functioning robots~\cite{iso10218part1,iso10218part2}.
Introducing new robots demands comprehensive re-consideration of the interaction between humans and robots and re-design of the workspace.
This will further pose new safety and risk management challenges within the existing system.

Besides, some production systems are non-stop, e.g., the final assembly line in the automotive industry~\cite{chen2015robotic}.
This indicates the necessity of automation systems handling the assembly during the movement of products.
Thus, it is inevitable that the development of automation systems in such scenarios will consider how to synchronize the robotic manipulators with the moving assembly line while executing the assembly operations.

Moreover, multiple variants of products are commonly produced on the same production line.
This increases the complexity of the design of automation control systems.
The multiple variants also challenge the adaptiveness and agility of automation systems regarding different product variants.
Besides, the requirement for automation in actual production varies among different sub-sections within an industry.
For example, in the automotive industry, passenger vehicle production differs from that of heavy vehicles in terms of the required production rate and assembly environment.
Diverse production requirements demand heterogeneous automation solutions and increase the workload of designing automation solutions.

In addition, a never-ending challenge in manufacturing is fulfilling the demand on takt time and maintaining and improving productivity.
This everlasting task requires automation systems to operate reliably within a limited time.

\subsection{Automation challenges in wire harness assembly}

Automated assembly has been adopted in the automotive industry for years to fulfill the continuously increasing production requirements~\cite{michalos2010automotive,heydaryan2018safety}.
However, among other assembly operations, most of the operations of wire harness assembly in current production remain manual and challenging to automate~\cite{salunkhe2023review}.

Challenges in the robotic manipulation of DLOs exist in automating wire harness assembly.
Like robotic DLO manipulation, automating the assembly of wire harnesses is challenging due to the flexibility of wire harnesses.
It is complicated to recognize the long and uneven shape, estimate the state of a deformable wire harness, and control the force for manipulation~\cite{koo2008development,zhang2023learning}.
Usually, it is also complex to design the moving paths of robots to avoid the formation of knots and entanglements that could block the assembly process or even break the harness~\cite{zhang2023learning}.
Besides the deformability of the cables, the complex structures and non-rigid materials of other wire harness components exacerbate the challenge of robotic manipulation of wire harnesses~\cite{sun2010robotic}.
The requirement from the actual production for extremely tight position accuracy in some assembly scenarios and the availability of precise contactless measurement to the state of the target wire in real-time further makes many proposals challenging to implement and unreliable in actual production~\cite{jiang2015robotized}.

Furthermore, wire harness assembly is more complex than generic robotic manipulation of DLOs.
Considering the tree-like structure of wire harnesses, wire harnesses consist of a bundle of DLOs (e.g., wires) and a group of rigid objects (e.g., connectors and clamps)~\cite{aguirre1994economic}.
With this, previous research has further considered wire harnesses as semi-deformable linear objects~\cite{zhou2020practical} or branched deformable linear objects~\cite{wnuk2021tracking}.
Hence, the interaction and constraint among different wire harness branches demand further investigation.

\subsection{Current manual assembly of automotive wire harnesses}
\label{subsec:operation}

Based on current work instructions at Volvo Car Corporation and empirical data collected during visual inspection of a production line in a car manufacturing plant, the current manual assembly of automotive wire harnesses in the passenger cabins of automobiles can be summarized into the following five procedures: 1) prepare, 2) transport, 3) untangle, 4) route, and 5) assemble.

\textbf{Prepare}
Initially, the wire harnesses arrive at the assembly station tied and packed in a plastic bag or box.
The wire harness is too stiff to manipulate by human operators in later assembly processes manually.
Therefore, the pack of wire harnesses is warmed in an oven first to soften them so that human operators can manipulate them manually.
Since the wire harnesses arrive as tied and packed in sequence, and the positions of wire harness delivery and the oven are fixed, this procedure could be automated by implementing automated conveyors and conventional industrial robots.

\textbf{Transport}
After getting warmed, the wire harness is transported by a lifting machine operated by a human operator and dumped in the cabin.
In this stage, the wire harness remains tied in a whole chunk.
Therefore, conventional industrial robots could be programmed to pick up the warmed wire harnesses from the oven and release them into the vehicle body.

\textbf{Untangle}
After placing the wire harness, human operators bend into the car body to untie and untangle the wire harness manually.
To automate this procedure, robots require fast and accurate perception of various properties of the wire harnesses, such as shape and topology, before they start manipulating them.
During manipulation, robots need to track the deformation and adapt their control flexibly and promptly, which requires more intelligent robotic perception and control capabilities.

\textbf{Route}
After disentangling, a wire harness is routed in the car body manually so that different branches of a wire harness reach the mating area based on functionality.
Similar to automating the untangling procedure, robots require the capability to manipulate DLOs before robotizing this procedure.

\textbf{Assemble}
Lastly, human operators manually mate the clamps and connectors of the wire harness to the counterparts in the car.
Robotizing this procedure requires robotic perception of the positions and orientations of manipulating objects and more advanced robotic planning and control capability.

\section{Methodology}
\label{sec:method}

The systematic literature review is important for comprehensively understanding a subject's state of the art and identifying the gaps requiring future research~\cite{kitchenham2004procedures,rowley2004conducting,denyer2009producing}.
The systematic literature review in this study followed the methodology for planning and conducting a review suggested by \cite{kitchenham2004procedures}.
The co-authors of this article also followed the methodology of investigator triangulation~\cite{denzin2009research}, peer debriefing~\cite{janesick2015peer}, and expert review~\cite{tracey2009design} to strengthen the research quality.
A review protocol was developed first to ensure a systematic and reproducible review method, as shown in Table~\ref{tab:protocol}.

\begin{table*}[htbp]
    \caption{Review protocol for a systematic literature review on computer vision applications in the robotized wire harness assembly.}
    \centering
    \resizebox{\linewidth}{!}{
        \begin{tabular}{ll}
            \toprule
            Review criteria    &                                                                                                          \\
            \midrule
            Database           & \textit{Scopus}                                                                                          \\
            Search string      & (wir* OR cabl*) AND (harness* OR bundl*) AND assembl*                                                    \\
            Search field       & Article title, Abstract, Keywords                                                                        \\
            Subject area       & Engineering; Computer Science; Multidisciplinary; Business, Management and Accounting; Decision Sciences \\
            Article language   & English                                                                                                  \\
            Inclusion criteria & Proposing computer vision-based algorithm and/or technology for robotized wire harness assembly          \\
            Exclusion criteria & Not about wire harnesses; not about robotic assembly; about the manufacturing of wire harnesses          \\
            Search date        & September 6, 2023                                                                                        \\
            \bottomrule
        \end{tabular}
    }
    \label{tab:protocol}
\end{table*}

\subsection{Literature search}

The literature search was conducted on the \textit{Scopus} database.
\textit{Scopus} was selected as the online database for searching scientific peer-reviewed articles, considering it a de facto reference standard for the engineering community~\cite{chou2012comparison} and its adequate coverage of publications provided by various publishers.
It is preferred because of its higher inclusiveness in terms of contributions over the \textit{Web of Science} database~\cite{harzing2016google} and its higher reliability of collected peer-reviewed sources over the \textit{Google Scholar} database~\cite{martin2021google}.

Three researchers deliberated the keywords and the string for searching and the inclusion and exclusion criteria for scrutinizing to ensure the identification of as many relevant articles as possible.
After several search trials with different combinations of keywords, the following string was defined for the search within the field of \textit{Article title}, \textit{Abstract}, \textit{Keywords} on \textit{Scopus}: (wir* OR cabl*) AND (harness* OR bundl*) AND assembl*.
The asterisk character was used in the string to retrieve more results based on term variations.
The words ``cable'' and ``bundle'' were included as synonyms for ``wire'' and ``harness''.

The initial search returned a set of $1022$ articles.
A filter on the subject area was conducted on the initial search to limit the subject areas to \textit{Engineering}, \textit{Computer Science}, \textit{Decision Sciences}, \textit{Multidisciplinary}, and \textit{Business, Management and Accounting}.
The intention was to exclude studies referring to irrelevant subjects, which reduced the number of items to $728$.
The language of the article was also limited to English.
In addition, no filter regarding the year of publication was implemented, i.e., all past research works were kept for screening.
Finally, $662$ articles were identified on September 6, 2023.

\subsection{Literature selection}

\begin{figure}[htbp]
    \centering
    \includegraphics[width=\linewidth]{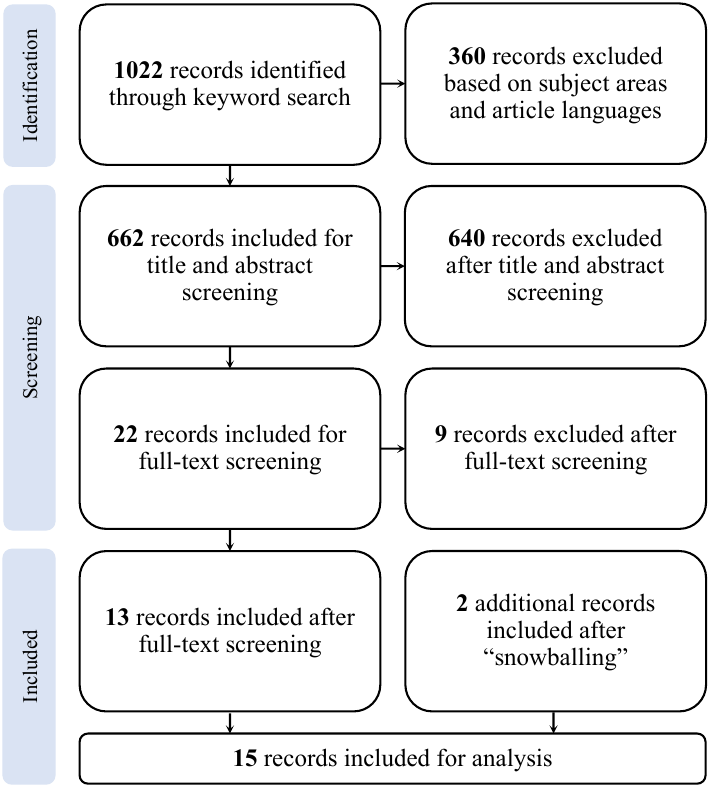}
    \caption{PRISMA~\cite{page2021prisma} flow diagram of review process. \textbf{Color should be used for this figure in print and size should be calibrated for the camera-ready version (This sentence will be removed for the final submission).}}
    \label{fig:prisma}
\end{figure}

After the literature search, a two-step screening was conducted by three researchers regarding the inclusion and exclusion criteria shown in Table~\ref{tab:protocol} to select the literature for later data synthesis and analysis.
The article selection process is reported following the Preferred Reporting Items for Systematic Reviews and Meta-Analyses (PRISMA)~\cite{page2021prisma} in Figure~\ref{fig:prisma}.

Following the selection criteria in Table~\ref{tab:protocol}, a study would be excluded if it did not regard wire harnesses, as described in Section~\ref{subsec:why-needed}, as the object of interest in the study.
A study would also be excluded if no robot system was involved in the proposed solution for wire harness assembly.
Then, as clarified at the end of Section~\ref{sec:intro}, articles that proposed computer vision-based solutions for robotizing wire harness assembly would be qualified for the analysis in this study, while articles presenting studies that focused on manufacturing wire harnesses would be excluded from the analysis.

In the first round of screening, the title and abstract of the $662$ articles were initially examined by three researchers independently based on the inclusion and exclusion criteria to minimize subjective bias during screening.
After screening individually, all three researchers synchronized their opinions and agreed on disagreements.
Firstly, $452$ articles were excluded because they were about something other than the wire harnesses focused in this review.
Then, $93$ articles were excluded because no robotic assembly was involved.
Furthermore, $83$ articles were excluded because they addressed the robotized assembly process in manufacturing wire harnesses.
Lastly, $12$ articles were excluded because they addressed robotized wire harness assembly without proposing vision-based solutions.
After the first screening on title and abstract, $640$ articles were excluded; thus, $22$ articles qualified, whose full texts were downloaded and evaluated in the second round of screening.

In the second round of screening, the full texts of downloaded articles were scrutinized meticulously according to the inclusion and exclusion criteria to filter the articles for later analysis, which resulted in $9$ records being further excluded due to failing to fulfill different criteria.
Hence, after the second round of screening, there were $13$ articles left for full-text analysis.

Moreover, ``snowballing''~\cite{greenhalgh2005effectiveness}, including \textit{reference tracking} and \textit{citation tracking}, was implemented on the selected $13$ articles for analysis on \textit{Scopus} to identify other relevant articles missed in the original search, which returned $2$ more articles.
Finally, $15$ peer-reviewed scientific articles were identified for further extensive, qualitative, descriptive, and quantitative data synthesis and analysis.

\subsection{Research quality and limitations}
\label{subsec:research-quality}

There are two significant perspectives for research quality evaluation: validity and reliability~\cite{adcock2001measurement,creswell2014research,mohajan2017two,patino2018internal,surucu2020validity}.
Validity suggests 1) to what extent the study prevents systematic errors appearing in designing and conducting the research (internal validity) and 2) to what extent the research findings can be generalized and applied to similar populations outside the study (external validity)~\cite{kitchenham2004procedures,creswell2014research,patino2018internal}.
Reliability measures the consistency of the usage of research approaches in similar studies~\cite{gibbs2007analyzing}.
A publicly available and widely adopted systematic review methodology, \cite{kitchenham2004procedures}, was followed to strengthen the quality of this systematic review.
This systematic review also referred to the Database of Abstracts of Reviews of Effects (DARE) criteria\footnote{\url{https://www.crd.york.ac.uk/CRDWeb/}} from the University of York, Centre for Reviews and Dissemination~\cite{petticrew1999quality} as a guideline to self-check the quality of this systematic review.
The DARE criteria~\cite{petticrew1999quality} qualifies a systematic review that addresses the first three criteria and at least one of the fourth and the fifth criteria:

\begin{enumerate}
    \item The existence of appropriate inclusion/exclusion criteria description
    \item The adequate search on relevant literature
    \item The existence of study synthesis
    \item The existence of quality assessment on the included studies
    \item The existence of sufficient details of the included studies in the review
\end{enumerate}

Researcher bias is generally unavoidable in qualitative studies.
Researchers' backgrounds and prior knowledge inevitably influence the objectiveness of the research.
However, this systematic review required prior computer vision, robotics, automation, and production knowledge.
It was used as a positive driver to design the research methods, analyze and interpret the findings, and propose high-quality future research directions.
Nevertheless, as suggested in \cite{kitchenham2004procedures}, a review protocol, including the methods for literature searching, screening, and selection, was calibrated beforehand to minimize the impact of researcher bias and enhance the objectiveness of this review.
Investigator triangulation~\cite{denzin2009research}, peer debriefing~\cite{janesick2015peer}, and expert review~\cite{tracey2009design} were also adopted through the study, where experts in relevant subjects from academia and industry (some of them as co-authors) were involved in calibrating the research methods and cross-validate the findings and interpretation.

The database is another crucial aspect of systematic review and literature search.
There have already been various bibliometric databases, e.g., \textit{Web of Science}, \textit{Scopus}, \textit{Google Scholar}, \textit{Microsoft Academic}, and \textit{Dimensions}~\cite{martin2021google}, which makes traversing all databases for a systematic review arduous.
Thus, database selection is necessary to improve operability while guaranteeing the adequacy of the search for relevant literature.
Though gaps exist among databases' coverage, \textit{Scopus} was selected in this systematic review, considering the consistency within the engineering community~\cite{chou2012comparison}, its higher inclusiveness over \textit{Web of Science}~\cite{harzing2016google}, and its higher reliable collection of peer-reviewed articles over \textit{Google Scholar}~\cite{martin2021google}.
This systematic review also adopted ``snowballing''~\cite{greenhalgh2005effectiveness} to mitigate the impact and complement the search result.

The identified studies are summarized, synthesized, and analyzed in the following sections to fulfill the rest of DARE criteria~\cite{petticrew1999quality}.

\section{Results}
\label{sec:result}

After searching and screening following the methodology described in Section~\ref{sec:method}, $15$ articles were identified for analysis, including $5$ articles and $10$ conference papers regarding the document type.
The contributions of these $15$ articles regarding computer vision applications are summarized in Table~\ref{tab:paper_summary}.

\begin{table*}[htbp]
    \caption{Identified works and their contributions regarding computer vision applications.}
    \centering
    \resizebox{\linewidth}{!}{
        \begin{tabular}{cccp{13cm}}
            \toprule
            Article                      & Year & Component        & Contributions in the aspect of computer vision applications                                                                                                                                                                                                                                                                                                                                                                                                                                                                                                                                              \\
            \midrule
            \cite{koo2008development}    & 2008 & Clamp            & This paper proposed to use two stereo vision systems mounted on each end effector of two robot arms to recognize the designed markers on the designed cubic clamp covers using Scale Invariant Feature Transform (SIFT)~\cite{lowe1999object,lowe2004distinctive}. The experimental results indicated that the vision system can provide enough precision for gripping clamp covers.                                                                                                                                                                                                                     \\
            \midrule
            \cite{di2009hybrid}          & 2009 & Connector        & This paper proposed to use one charge-coupled-device (CCD) camera for connector grasping error detection and quality control based on basic pattern matching.                                                                                                                                                                                                                                                                                                                                                                                                                                                                    \\
            \midrule
            \cite{jiang2010robotized}    & 2010 & Clamp            & This paper improved \cite{koo2008development} by using three stereo vision systems mounted on each end effector of three robot arms and ten fixed cameras surrounding the work cell to recognize the ARToolKit~\cite{kato1999marker} markers on the designed cylinder-like-shape clamp covers.                                                                                                                                                                                                                                                                                                 \\
            \midrule
            \cite{sun2010robotic}        & 2010 & Connector        & This paper improved \cite{di2009hybrid} by using two mutually perpendicular CCD cameras to detect magnitudes of tilt angles and horizontal displacements from each side by pattern matching.                                                                                                                                                                                                                                                                                                                                                                                                             \\
            \midrule
            \cite{jiang2011robotized}    & 2011 & Clamp            & This paper extended \cite{jiang2010robotized} with more details and discussions.                                                                                                                                                                                                                                                                                                                                                                                                                                                                                                                         \\
            \midrule
            \cite{di2012vision}          & 2012 & Connector        & This paper focused on the monitoring task for the connector mating process. Two mutually perpendicular cameras were used to observe the relative and online motions between the two connectors based on pattern matching.                                                                                                                                                                                                                                                                                                                                                                                \\
            \midrule
            \cite{tamada2013high}        & 2013 & Connector        & This paper proposed to use a high-speed vision system to acquire the categories, orientations, and positions of connectors at a frame rate of $500$ frames per second (FPS) via connector corner detection.                                                                                                                                                                                                                                                                                                                                                                                                                  \\
            \midrule
            \cite{jiang2015robotized}    & 2015 & Clamp            & This paper proposed to use a wrist camera on the right robot arm to recognize the ARToolKit~\cite{kato1999marker} markers on the clamp cover, whose position was estimated beforehand based on tracing trajectory.                                                                                                                                                                                                                                                                                                                                                                             \\
            \midrule
            \cite{song2017electric}      & 2017 & Connector        & This paper proposed a method for monitoring the mating process of electric connectors. A hand-eye camera was used for locating the connector headers via visual servoing with markers.                                                                                                                                                                                                                                                                                                                                                                                                                   \\
            \midrule
            \cite{yumbla2020preliminary} & 2020 & Connector        & This paper proposed using an red-green-blue-depth (RGB-D) camera for the recognition of plug-in cable connectors based on image processing. The positions of connectors were detected based on red-green-blue (RGB) images. The three-dimensional (3D) information of the detected connectors were then acquired by registering the depth information to the RGB-based detection results.                                                                                                                                                                                                                                                        \\
            \midrule
            \cite{zhou2020practical}     & 2020 & Connector        & This paper proposed to acquire the position and orientation of a connector via learning-based rough locating and shape-based fine positioning. The proposal used three cameras: 1) a fixed global camera for rough locating; and 2) a hand-eye camera per robot arm (two robot arms in total) for fine positioning.                                                                                                                                                                                                                                                                                      \\
            \midrule
            \cite{kicki2021tell}         & 2021 & Wire             & This paper focused on the interpretable classification of wire harness branches. The interpretability was visualized using saliency maps based on class activation mapping (CAM)~\cite{zhou2016learning}. The experimental results demonstrated the best classification based on a late prediction fusion~\cite{bednarek2020robustness} of RGB and depth modalities; the deteriorating performance with the network pre-trained on the inpainting task; and the positive effect of elastic transform for data augmentation. The saliency maps promoted the interpretability of the experimental results. \\
            \midrule
            \cite{guo2022visual}         & 2022 & Wire             & This paper proposed a multi-branch wire harness object recognition with segmentation and estimation based on point clouds acquired using an RGB-D camera.                                                                                                                                                                                                                                                                                                                                                                                                                                                \\
            \midrule
            \cite{zhang2023learning}     & 2023 & Wire             & This paper explored the industrial bin-picking problem on wire harnesses. The study proposed learning a bin-picking policy to infer an optimal grasp and a post-grasping action based on a top-down depth image of the cluttered wire harnesses captured by a Photoneo PhoXi 3D scanner M. This paper also suggested visual noise and heavy occlusion as two major challenges leading to failure results.                                                                                                                                                                                                \\
            \midrule
            \cite{zagar2023copy}         & 2023 & Wire harness bag & This paper focused on RGB-based deformable wire harness bag segmentation. A bag instance segmentation dataset generation pipeline based on the \textit{copy-and-paste} technique~\cite{dwibedi2017cut,ghiasi2021simple,wang2022cp2} and geometric and photometric image data augmentation techniques~\cite{shorten2019survey} was proposed to address the lack of annotated datasets of task-specific objects of interest.
            \\
            \bottomrule
        \end{tabular}
    }
    \label{tab:paper_summary}
\end{table*}

\begin{figure}[htbp]
    \centering
    \includegraphics[width=\linewidth]{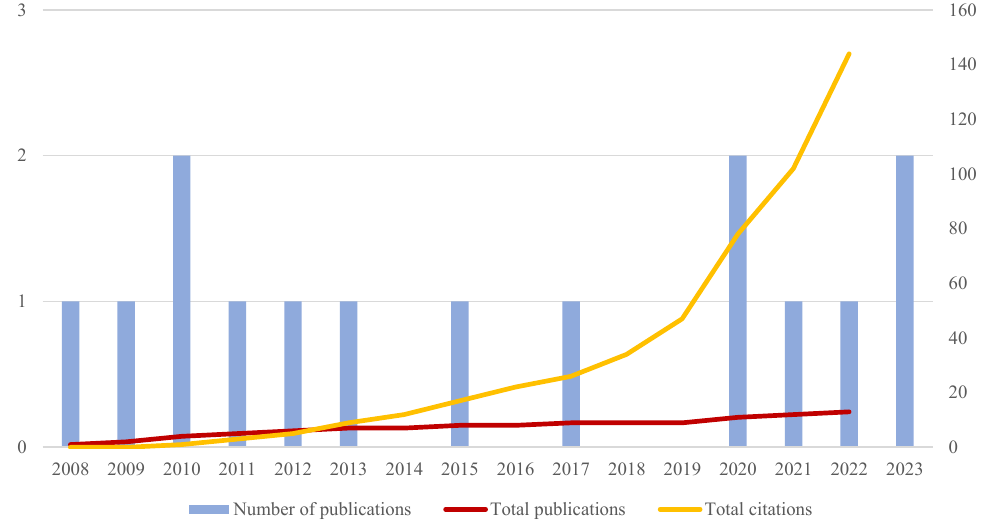}
    \caption{
    The number of publications per year and the total number of publications and citations by each year (data retrieved from \textit{Scopus} on September 6, 2023).
    The total number of publications and citations for the year 2023 are excluded considering the incomplete statistics at the time of literature searching.
    \textbf{Color should be used for this figure in print and size should be calibrated for the camera-ready version (This sentence will be removed for the final submission).}
    }
    \label{fig:stat_year}
\end{figure}

Figure~\ref{fig:stat_year} illustrates the development of research in computer vision applications in the robotized wire harness assembly throughout the past years regarding the total number of publications and citations.
The data was retrieved from \textit{Scopus} on September 6, 2023.
The statistics indicate a persistent long-term and increasing effort to facilitate the robotized wire harness assembly with vision systems.

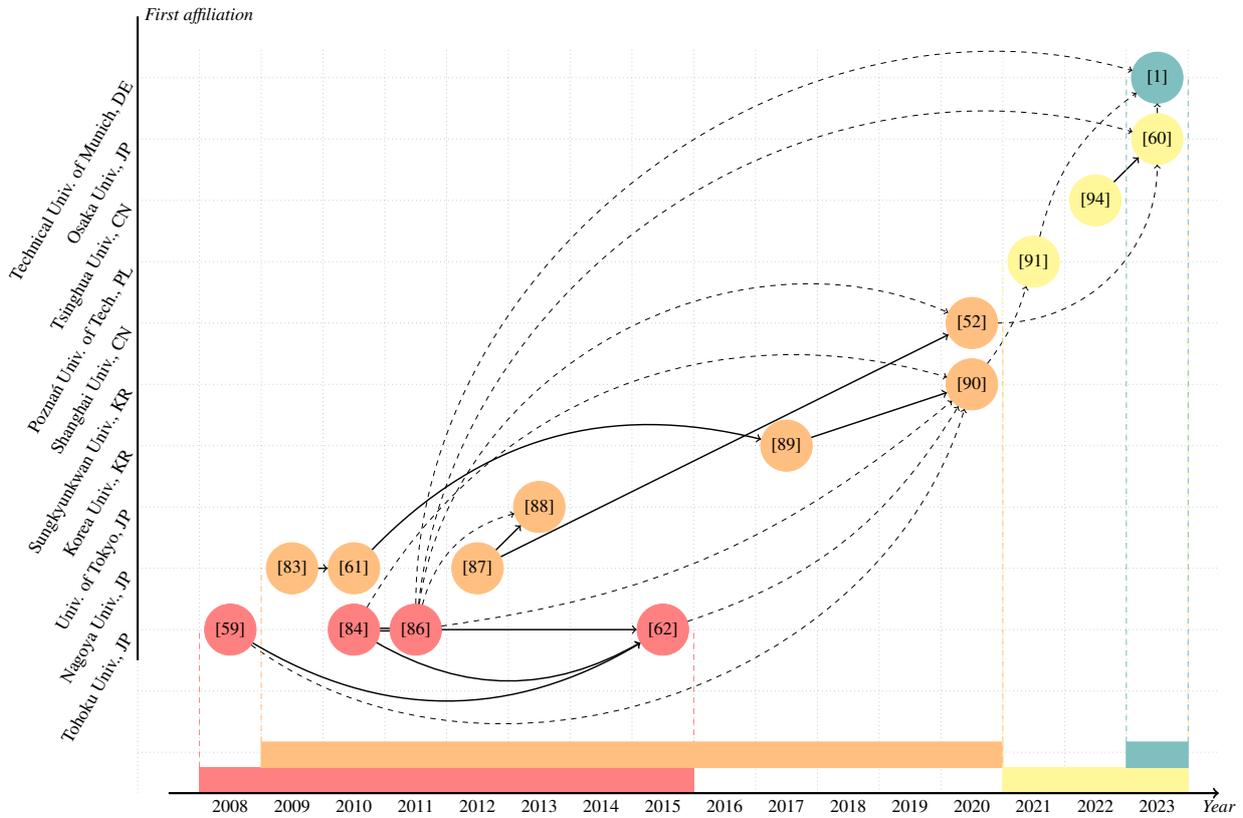
\begin{figure*}[htb]
    \centering
    \resizebox{\linewidth}{!}{
        \begin{tikzpicture}
            \node[draw=red!50, circle, fill=red!50, minimum size=10mm] (A) at (1.5*1.2,1*1.2) {\cite{koo2008development}};
            \node[draw=red!50, circle, fill=red!50, minimum size=10mm] (B) at (3.5*1.2,1*1.2) {\cite{jiang2010robotized}};
            \node[draw=red!50, circle, fill=red!50, minimum size=10mm] (C) at (4.5*1.2,1*1.2) {\cite{jiang2011robotized}};
            \node[draw=red!50, circle, fill=red!50, minimum size=10mm] (D) at (8.5*1.2,1*1.2) {\cite{jiang2015robotized}};
            \node[draw=orange!50, circle, fill=orange!50, minimum size=10mm] (E) at (2.5*1.2,2*1.2) {\cite{di2009hybrid}};
            \node[draw=orange!50, circle, fill=orange!50, minimum size=10mm] (F) at (3.5*1.2,2*1.2) {\cite{sun2010robotic}};
            \node[draw=orange!50, circle, fill=orange!50, minimum size=10mm] (G) at (5.5*1.2,2*1.2) {\cite{di2012vision}};
            \node[draw=orange!50, circle, fill=orange!50, minimum size=10mm] (H) at (6.5*1.2,3*1.2) {\cite{tamada2013high}};
            \node[draw=orange!50, circle, fill=orange!50, minimum size=10mm] (I) at (10.5*1.2,4*1.2) {\cite{song2017electric}};
            \node[draw=orange!50, circle, fill=orange!50, minimum size=10mm] (J) at (13.5*1.2,5*1.2) {\cite{yumbla2020preliminary}};
            \node[draw=orange!50, circle, fill=orange!50, minimum size=10mm] (K) at (13.5*1.2,6*1.2) {\cite{zhou2020practical}};
            \node[draw=yellow!50, circle, fill=yellow!50, minimum size=10mm] (L) at (14.5*1.2,7*1.2) {\cite{kicki2021tell}};
            \node[draw=yellow!50, circle, fill=yellow!50, minimum size=10mm] (M) at (15.5*1.2,8*1.2) {\cite{guo2022visual}};
            \node[draw=yellow!50, circle, fill=yellow!50, minimum size=10mm] (N) at (16.5*1.2,9*1.2) {\cite{zhang2023learning}};
            \node[draw=teal!50, circle, fill=teal!50, minimum size=10mm] (O) at (16.5*1.2,10*1.2) {\cite{zagar2023copy}};

            \draw[->, black, thick] (A) to [bend right=31] (D);
            \draw[->, black, dashed] (A) to [bend right=55] (J);
            \draw[double, black, thick] (B) to (C);
            \draw[->, black, thick] (B) to [bend right=30] (D);
            \draw[->, black, dashed] (B) to [bend left=38] (J);
            \draw[->, black, thick] (C) to (D);
            \draw[->, black, dashed] (C) to [bend left] (H);
            \draw[->, black, dashed] (C) to [bend right=16] (J);
            \draw[->, black, dashed] (C) to [bend left=53] (K);
            \draw[->, black, dashed] (C) to [bend left=50] (N);
            \draw[->, black, dashed] (D) to [bend right=20] (J);
            \draw[->, black, thick] (E) to (F);
            \draw[->, black, thick] (F) to [bend left] (I);
            \draw[->, black, thick] (G) to (H);
            \draw[->, black, thick] (G) to (K);
            \draw[->, black, thick] (I) to (J);
            \draw[->, black, dashed] (J) to [bend right=10] (L);
            \draw[->, black, dashed] (K) to [bend right=45] (N);
            \draw[->, black, thick] (M) to (N);
            \draw[->, black, dashed] (C) to [bend left=53] (O);
            \draw[->, black, dashed] (L) to [bend left=20] (O);
            \draw[->, black, dashed] (N) to (O);

            \draw[red!50, fill=red!50] (1*1.2,-2) rectangle (9*1.2,-1.5);
            \draw[red!50, dashed] (1*1.2,-1.5) -- (1*1.2,1*1.2);
            \draw[red!50, dashed] (9*1.2,-1.5) -- (9*1.2,1*1.2);
            \draw[orange!50, fill=orange!50] (2*1.2,-1.5) rectangle (14*1.2,-1);
            \draw[orange!50, dashed] (2*1.2,-1) -- (2*1.2,2*1.2);
            \draw[orange!50, dashed] (14*1.2,-1) -- (14*1.2,6*1.2);
            \draw[yellow!50, fill=yellow!50] (14*1.2,-2) rectangle (17*1.2,-1.5);
            \draw[yellow!50, dashed] (14*1.2,-1.5) -- (14*1.2,7*1.2);
            \draw[yellow!50, dashed] (17*1.2,-1.5) -- (17*1.2,9*1.2);
            \draw[teal!50, fill=teal!50] (16*1.2,-1.5) rectangle (17*1.2,-1);
            \draw[teal!50, dashed] (16*1.2,-1) -- (16*1.2,10*1.2);
            \draw[teal!50, dashed] (17*1.2,-1) -- (17*1.2,10*1.2);

            \draw[->,very thick] (0.5*1.2,-2) to (17.5*1.2,-2) node[below] {\textit{Year}};
            \draw[-,very thick] (0,0.5*1.2) to (0,11*1.2) node[right] {\textit{First\ affiliation}};
            \begin{scope}[on background layer]
                \draw[style=help lines, step=1*1.2cm, dotted] (0,-2) grid (17.5*1.2,10.5*1.2);
            \end{scope}

            \foreach \x [evaluate=\x as \year using int(\x+2007)] in {1,...,16} \node[below] at (\x*1.2+0.5*1.2,-2) {\year};
            \node[left, rotate=60] at (-0.1,1*1.2) {Tohoku Univ., JP};
            \node[left, rotate=60] at (-0.1,2*1.2) {Nagoya Univ., JP};
            \node[left, rotate=60] at (-0.1,3*1.2) {Univ. of Tokyo, JP};
            \node[left, rotate=60] at (-0.1,4*1.2) {Korea Univ., KR};
            \node[left, rotate=60] at (-0.1,5*1.2) {Sungkyunkwan Univ., KR};
            \node[left, rotate=60] at (-0.1,6*1.2) {Shanghai Univ., CN};
            \node[left, rotate=60] at (-0.1,7*1.2) {Pozna{\'n} Univ. of Tech., PL};
            \node[left, rotate=60] at (-0.1,8*1.2) {Tsinghua Univ., CN};
            \node[left, rotate=60] at (-0.1,9*1.2) {Osaka Univ., JP};
            \node[left, rotate=60] at (-0.1,10*1.2) {Technical Univ. of Munich, DE};

        \end{tikzpicture}
    }
    \caption{
    The citation map among the identified studies.
    The red, orange, yellow, and teal circles and bars represent previous studies focusing on clamps, connectors, wires, and wire harness bags.
    Solid and dashed arrows represent the citation between two studies discussing the same and different wire harness components.
    The double solid line represents the extension of the core content.
    (Univ. = University; Tech. = Technology)
    \textbf{Color should be used for this figure in print and size should be calibrated for the camera-ready version (This sentence will be removed for the final submission).}
    }
    \label{fig:citation_map}
\end{figure*}

Figure~\ref{fig:citation_map} presents the citation relationships among the identified $15$ studies.
These relationships illustrate that the research in vision-based robotized wire harness assembly began with the recognition and manipulation of wire harness components, such as clamps and connectors.
More recent studies have initiated new directions involving understanding the overall structure of wire harnesses.
However, the research regarding different components of wire harnesses ceased at different times.

In Figure~\ref{fig:citation_map}, each study is represented by a colorized circle in the two-dimensional (2D) Cartesian coordinate system regarding the year of publication, the first affiliation of the paper, and the wire harness component in focus.
The red, orange, yellow, and teal circles represent previous studies focusing on clamps, connectors, wires, and wire harness bags.
Correspondingly, the red, orange, yellow, and teal bars by the $x$-axis indicate the period of previous studies focusing on clamps, connectors, wires, and wire harness bags, respectively.
Solid and dashed arrows visualize the citation relationships.
These arrows represent the citation between two studies discussing the same and different wire harness components.
Each arrow represents a citation-reference relationship between two previous studies.
For example, ``A''$\rightarrow$``B'' means that ``A'' was cited by ``B''.
Solid arrows represent the citation between two studies focusing on the same component.
Dashed arrows represent the citation between two studies focusing on different components.
The double solid line between \cite{jiang2010robotized} and \cite{jiang2011robotized} represents the extension on details and discussions in \cite{jiang2011robotized} from \cite{jiang2010robotized}.

\subsection{Research regarding wire harness components}

The studies identified for this systematic literature review were experimental studies in laboratory settings.
This review categorizes these studies into four main groups regarding the wire harness component in focus: 1) for clamp manipulation ($4$ studies); 2) for the mating of connectors ($7$ studies); 3) for wire harness recognition ($3$ studies); and 4) for wire harness bag segmentation ($1$ study).

\subsubsection{Clamp manipulation}

Clamps are bound on wire harnesses for attaching wire harnesses to target positions on the final product.
Four reviewed articles addressed the manipulation of clamps on wire harnesses achieved with computer vision-based clamp recognition, as summarized in Table~\ref{tab:clamp}.

\begin{table}[htbp]
    \caption{
    Computer vision-based clamp recognition for clamp manipulation.
    (LC = location of cameras; TV = type of vision; NC = number of cameras)
    }
    \centering
        \begin{tabular}{ccccc}
            \toprule
            LC & Article                                      & TV & NC & Method                                         \\
            \midrule
            End effector        & \cite{koo2008development}                    & Stereo         & $4$               & SIFT~\cite{lowe1999object,lowe2004distinctive} \\
                                & \cite{jiang2010robotized,jiang2011robotized} & Stereo         & $6$               & ARToolKit~\cite{kato1999marker}      \\
                                & \cite{jiang2015robotized}                    & Monocular      & $1$               & ARToolKit~\cite{kato1999marker}      \\
            Work cell           & \cite{jiang2010robotized,jiang2011robotized} & Monocular      & $10$              & ARToolKit~\cite{kato1999marker}      \\
            \bottomrule
        \end{tabular}
    \label{tab:clamp}
\end{table}

This group of articles focused on mating wire harnesses onto a car body, e.g., inserting the clamps on a wire harness into an instrument panel frame.
In these proposals, clamps were recognized using CCD camera~\cite{koo2008development,jiang2010robotized,jiang2011robotized} and complementary-metal–oxide–semiconductor (CMOS) camera~\cite{jiang2015robotized}.
With the recognition results, the gripper or end effector on a robot arm could reach the clamps for further manipulations.
All four studies implemented hand-eye vision systems by mounting cameras on the end effectors of robot arms~\cite{koo2008development,jiang2010robotized,jiang2011robotized,jiang2015robotized}.
Two of them also adopted global vision systems with multiple cameras fixed around the operation area~\cite{jiang2010robotized,jiang2011robotized}.

Although the clamps can be regarded as rigid objects, their relatively small sizes and complex shapes make them difficult to grasp by a robot gripper directly~\cite{koo2008development}.
The complexity of control program development and the insufficiency of the payload makes multi-finger robot hands challenging to adapt in practical applications~\cite{koo2008development}.
Thus, \cite{koo2008development} proposed to install cubic clamp covers to facilitate the manipulation of clamps with small sizes and complex shapes.
Then, \cite{koo2008development} introduced stereo vision systems mounted on the end effectors of two robot arms to recognize the designed markers on the clamp covers based on SIFT~\cite{lowe1999object,lowe2004distinctive}.
The stereo vision system in \cite{koo2008development} contained two CCD cameras with different focal lengths for observing far and near objects, respectively.
Based on the recognition results, the robot arms moved to the clamp covers for further manipulation.
The experimental results in \cite{koo2008development} qualitatively indicated that the implemented vision system could visually recognize clamp covers with enough precision for robots to grip clamp covers~\cite{koo2008development}.

Later, \cite{jiang2010robotized} and \cite{jiang2011robotized} made further improvements to \cite{koo2008development} by:
1) deploying one more robot arm;
2) using newly designed cylinder-like-shape clamp covers with more markers from ARToolKit~\cite{kato1999marker} for recognition;
3) implementing a global vision system (ten fixed cameras surrounding the work frame from various directions) besides the three hand-eye vision systems for clamp cover detection to address the occlusion problem on clamps (this increased the number of cameras to sixteen);
and 4) adding a laser head on the end effector of each robot arm in case precise measurement of a wire segment was demanded.

Instead of directly detecting the clamp covers on a wire harness visually, \cite{jiang2015robotized} proposed to adopt a mechanical tracing operation to identify the position of clamp covers on a wire harness first.
Then, the markers on the clamp cover were recognized with one wrist CMOS camera (Point Grey Firefly MV) on the right robot arm to estimate the pose of the identified clamp cover so that the robot could manipulate it in later procedures~\cite{jiang2015robotized}.

\subsubsection{Mating of connectors}

Mating of connectors is critical in the final assembly of wire harnesses in terms of function and quality.
Table~\ref{tab:connector} summarizes the seven articles focusing on different stages of the mating of connectors, which can be grouped into three sub-procedures: 1) pre-assembly, 2) mating, and 3) post-assembly.

\begin{table*}[htbp]
    \caption{
    Computer vision applications in articles addressing the mating of connectors.
    (TV = type of vision; TC = type of cameras; LC = location of cameras; NC = number of cameras; GF = globally fixed; HE = hand-eye; 2.5D = two-and-a-half dimensional)
    }
    \centering
    \resizebox{\linewidth}{!}{
        \begin{tabular}{cccccccc}
            \toprule
            \multirow{2}{*}{Process} & \multirow{2}{*}{Purpose}    & \multicolumn{2}{c}{TV}   & \multirow{2}{*}{TC}          & \multirow{2}{*}{LC} & \multirow{2}{*}{NC} & \multirow{2}{*}{Method}                                         \\
            \cmidrule{3-4}
                                     &                             & 2D                       & 2.5D                         &                     &                     &                         &                                       \\
            \midrule
            Pre-assembly             & Connector detection         & \cite{tamada2013high}    & -                            & MC1362              & GF                  & $1$                     & Image processing                      \\
                                     &                             & -                        & \cite{yumbla2020preliminary} & RealSense D435      & HE                  & $1$                     & Image processing                      \\
                                     &                             & \cite{zhou2020practical} & -                            & Industrial cameras  & GF+HE               & $1$+$2$                 & Learning- and model-based             \\
                                     & Pose estimation             & \cite{zhou2020practical} & -                            & Industrial cameras  & GF+HE               & $1$+$2$                 & Learning- and model-based             \\
                                     & Fault detection             & \cite{di2009hybrid}      & -                            & In-Sight 5100       & GF                  & $1$                     & Pattern matching                      \\
                                     &                             & \cite{sun2010robotic}    & -                            & CCD cameras         & GF                  & $2$                     & Pattern matching                      \\
            Mating                   & Relative position detection & \cite{di2012vision}      & -                            & CCD cameras         & GF                  & $2$                     & Pattern matching                      \\
                                     &                             & \cite{tamada2013high}    & -                            & MC1362              & GF                  & $1$                     & Image processing                      \\
                                     &                             & \cite{song2017electric}  & -                            & FL2G-13S2C-C        & HE                  & $1$                     & Visual servoing with pattern matching \\
            Post-assembly            & Fault detection             & \cite{di2009hybrid}      & -                            & In-Sight 5100       & GF                  & $1$                     & Pattern matching                      \\
            \bottomrule
        \end{tabular}
    }
    \label{tab:connector}
\end{table*}

Pre-assembly is the initial process where the vision system acquires the necessary information to guide a robot's movement toward connectors and the later robotized mating operation.
There were five articles discussing various tasks in pre-assembly, including connector detection~\cite{tamada2013high,yumbla2020preliminary,zhou2020practical}, pose estimation~\cite{zhou2020practical}, and fault detection~\cite{di2009hybrid,sun2010robotic}.

Perceiving the 3D geometric information of connectors, e.g., position and orientation, is essential for robotized mating connectors on flexible wire harnesses.
\cite{tamada2013high} proposed to use a high-speed camera (EoSens series, MC1362, Mikrotron) fixed above the workbench to distinguish the types of connectors and acquire their positions and orientations by detecting the corners of the connectors at a frame rate of $500$ FPS.
Eight-bit grayscale images were captured and processed through a sequence of image processing methods from OpenCV library\footnote{\url{https://opencv.org}}.
The image processing included image smoothing, binarization, contour detection, and straight-line approximation to acquire the corners of each connector~\cite{tamada2013high}.
The recognized connector corners were further used for obtaining the orientation of each connector by calculating each connector's principal axis of inertia~\cite{tamada2013high}.
\cite{tamada2013high} processed 2D grayscale images only as they designed that the connectors handled by the robot hand were placed on a table surface to make the $z$-coordinate of every connector constant and the orientation of every connector various in the yaw direction only.

Targeted on obtaining the precise positions of the plug-in cable connectors on a workbench, \cite{yumbla2020preliminary} proposed to adopt an Intel RealSense D435 depth camera for the recognition process based on image processing.
The captured RGB images were first processed through a sequence of image processing methods from the OpenCV library, including color space conversion, color thresholding, and image moment calculation, to acquire connectors' 2D contours on $XY$-plane~\cite{yumbla2020preliminary}.
Then, the obtained 2D contours were mapped to the captured depth image, which was registered to the RGB image in advance, to obtain the 3D information of connectors on $z$-direction for further robotized manipulation~\cite{yumbla2020preliminary}.

A two-step connector detection algorithm was proposed in \cite{zhou2020practical} and verified with a dual-arm robot using three cameras (one fixed global camera and one hand-eye camera on each robot arm).
The first step in \cite{zhou2020practical} was a learning-based rough locating, where a 2D grayscale image of the wire harness captured by the fixed global camera was processed by MobileNet-SSD~\cite{howard2017mobilenets} to locate the connectors roughly.
Then, one of the hand-eye cameras on the robot arms reached the top of the located connector to capture images for a shape-based fine positioning to obtain the 6-degrees-of-freedom (6-DOF) pose of the detected connector for further manipulation based on computer-aided design (CAD) models and multi-view image matching~\cite{zhou2020practical}.

Previous studies have also discussed quality assurance in collecting connectors by the gripper on a robot arm.
\cite{di2009hybrid} proposed to use an In-Sight 5100 camera for vision-based connector grasping error detection and quality control to confirm that the connector was caught correctly and conveyed to the correct position.
The relative translational and rotational displacements between the gripper and the connector were examined based on basic pattern matching with image processing~\cite{di2009hybrid}.
The pattern matching algorithm processed a cross-shape pattern drawn on the gripper, a circle-shape pattern drawn on the female connector, and the intrinsic design of the male connector~\cite{di2009hybrid}.
Besides, \cite{di2009hybrid} implemented this fault detection after inserting the connectors to examine the operation result.
However, the experiment results of \cite{di2009hybrid} indicated the insufficient training samples and the only single camera used as the main reasons behind the misclassifications.
\cite{sun2010robotic} further improved \cite{di2009hybrid} by adding one more camera perpendicular to the first one to detect magnitudes of tilt angles and horizontal displacements on each side based on pattern matching.

After being collected by a gripper, the connector is transported and mated to the target counterpart.
Previous studies proposed vision-based solutions for guiding and monitoring the mating process~\cite{di2012vision,tamada2013high,song2017electric}.
\cite{di2012vision} focused on the monitoring task and implemented two mutually perpendicular cameras to observe all the relative and online motions between the two connectors by pattern matching, whose design was akin to the pattern matching algorithms adopted by \cite{di2009hybrid} and \cite{sun2010robotic}.
To locate the connector headers, \cite{song2017electric} adopted a hand-eye camera (FL2G-13S2C-C, Point Grey Research) and proposed to implement visual servoing with markers~\cite{chaumette2006visual} to track the headers of connectors.
In addition to connector detection in pre-assembly, \cite{tamada2013high} also proposed to use the high-speed vision system to monitor the mating process in real-time by detecting the distance between the in-hand and the target connector.

\subsubsection{Wire harness recognition}

Besides clamps and connectors, the wire part of a wire harness is also critical to address.
The recognition of the wire part can provide a better perception of the overall structure of a wire harness.
There are three studies discussing wire harness recognition, as listed in Table~\ref{tab:wire}, focusing on the interpretable classification of branches~\cite{kicki2021tell}, wire recognition~\cite{guo2022visual}, and grasp detection~\cite{zhang2023learning}, respectively.

\begin{table*}[htbp]
    \caption{Computer vision applications in articles for wire harness recognition. (TC = type of cameras; LC = location of cameras; NC = number of cameras; GF = globally fixed)}
    \centering
        \begin{tabular}{cccccc}
            \toprule
            Article                  & Purpose                      & TC               & LC & NC    & Method                                            \\
            \midrule
            \cite{kicki2021tell}     & Interpretable classification & RealSense D435   & GF & ($1$) & Deep learning-based                               \\
            \cite{guo2022visual}     & Visual recognition           & RGB-D            & -  & -     & Machine learning-based                            \\
            \cite{zhang2023learning} & Grasp detection              & Photoneo PhoXi M & GF & $1$   & Fast Graspability Evaluation~\cite{domae2014fast} \\
            \bottomrule
        \end{tabular}
    \label{tab:wire}
\end{table*}

\cite{kicki2021tell} focused on the interpretable classification of wire harness branches.
Specifically, \cite{kicki2021tell} proposed a dataset and several convolutional neural networks (CNN)~\cite{goodfellow2016deep} for classification based on different data modalities.
The proposed dataset contained RGB-D images of four branches of an automotive wire harness captured by an Intel RealSense D435 depth camera mounted $50$ cm above the ground level~\cite{kicki2021tell}.
The CNNs proposed in \cite{kicki2021tell} shared the same Downsample layer from ERFNet~\cite{romera2018erfnet}.
Data augmentation with elastic transformation and a network pre-trained with the inpainting task were also evaluated in \cite{kicki2021tell}.
The experiment results of \cite{kicki2021tell} on various input modalities demonstrated the best accuracy achieved by classifying with a sum of logits of models taking RGB data and depth information as input, respectively, following the late prediction fusion~\cite{bednarek2020robustness}.
Further experiments dealing with small datasets in \cite{kicki2021tell} indicated a significant drop in the performance with the classification network pre-trained on the inpainting task but a performance improvement with data augmentation using elastic transform.
\cite{kicki2021tell} also visualized the classification results using a saliency map based on class activation mapping (CAM)~\cite{zhou2016learning} to interpret the experimental results better visually.

\cite{guo2022visual} proposed a multi-branch wire harness object recognition with sequential segmentation and probabilistic estimation in aircraft assembly.
In the proposal~\cite{guo2022visual}, the raw point cloud data was first acquired using an RGB-D camera.
Then, the raw data was pre-processed to supervoxels by over-segmentation.
Based on the supervoxels, wires were segmented considering the Cartesian distance, color similarity, and bending continuity.
After segmentation, there were inevitable gaps in the segmentation result due to sheltering or occlusion.
The gaps were further remedied by estimation with Gaussian Mixture Model (GMM)~\cite{reynolds2009gaussian} to obtain the complete segmented point cloud of wires~\cite{guo2022visual}.

Focusing on the industrial bin-picking problem on wire harnesses, \cite{zhang2023learning} proposed learning a bin-picking policy to infer an optimal grasp and a post-grasping action from a top-down depth image of the cluttered wire harnesses.
The proposal enabled the system to prioritize grasping the untangled objects, avoid grasping at the wrong positions, and reason the extracting distance to reduce the execution time for a successful picking~\cite{zhang2023learning}.
The vision system in \cite{zhang2023learning} included a Photoneo PhoXi 3D scanner M fixed directly above the workbench.
Given a top-down depth image as an observation of the current entanglement situation, a model-free grasp detection based on Fast Graspability Evaluation (FGE)~\cite{domae2014fast} was adapted in \cite{zhang2023learning} to detect collision-free grasps.
Specifically, the solution outputted a set of grasp candidates ordered by their FGE scores~\cite{zhang2023learning}.
Then, \cite{zhang2023learning} trained an action success prediction module via a proposed active learning scheme to predict the success possibilities of the disentangling actions.
Specifically, \cite{zhang2023learning} encoded and processed the captured depth image, the set of grasp candidates, and the pre-defined action candidates using different CNNs.
The action-grasp pair to conduct the operation was then selected based on the FGE~\cite{domae2014fast} score and the action complexity~\cite{zhang2023learning}.
However, the experiment results of \cite{zhang2023learning} indicated visual noise and heavy occlusion as two significant challenges leading to failure pickings.

\subsubsection{Wire harness bags segmentation}

Manufacturers have started organizing the overall wire harness into multiple sub-groups with deformable bags to simplify the assembly operations~\cite{zagar2023copy}.
To promote automating wire harness assembly operations, \cite{zagar2023copy} inquired about RGB-based deformable wire harness bag segmentation.
\cite{zagar2023copy} identified the lack of an annotated dataset of specific objects of interest as a significant constraint in developing required vision systems.
Hence, \cite{zagar2023copy} proposed a dataset generation pipeline relying on minimal human effort.
Specifically, the pipeline adopted the \textit{copy-and-paste} technique~\cite{dwibedi2017cut,ghiasi2021simple,wang2022cp2} to generate images with diverse combinations of objects and backgrounds~\cite{zagar2023copy}.
A list of geometric and photometric image data augmentation techniques~\cite{shorten2019survey} was also adopted to address the insufficient data of task-specific objects of interest, the deformable wire harness bags~\cite{zagar2023copy}.

First, $56$ foreground images with a single instance of a wire harness bag in each were captured and manually annotated using an annotation tool, \textit{Segments.ai}\footnote{\url{https://segments.ai/}}, to collect bag instances.
Then, indoor scene images in Massachusetts Institute of Technology (MIT) indoor scenes dataset~\cite{quattoni2009recognizing} were selected as background images for good generalization in real-world scenarios, considering the expected application environment in factories or industrial plants.
High-Resolution Salient Object Detection (HRSOD) dataset~\cite{zeng2019towards} and a ``complex'' dataset collected following the data collection methodology in \cite{zanella2021auto} were also adopted for comparison in evaluation.
A set of geometric and photometric image data augmentation techniques~\cite{shorten2019survey} were implemented on foreground and background images, respectively, before implementing \textit{copy-and-paste}~\cite{dwibedi2017cut,ghiasi2021simple,wang2022cp2} to enable greater diversity and variance among data samples in the final overall dataset.
In implementing \textit{copy-and-paste}~\cite{dwibedi2017cut,ghiasi2021simple,wang2022cp2}, an image of a wire harness bag was randomly selected and pasted $n$ times at random locations on a background image selected randomly.
The number $n$ of foreground instances in each training image was between $1$ and $4$.
In total, $5000$ samples with a resolution of $640 \times 480$ pixels were obtained, with the $90\%$-$10\%$ training-validation split for training the segmentation model.

ResNet101~\cite{he2016deep}, SwinS~\cite{liu2021swin}, and ConvNeXtS~\cite{liu2022convnet} were selected as backbone architectures for the evaluation of the obtained dataset on the segmentation task of wire harness bags.
The trained models were evaluated with Dice coefficient and Intersection-over-Union (IoU) on a test set comprising $75$ accurately annotated images grouped into three scenarios ($25$ images each), including the laboratory scenario, the beginning of assembly operations, and during the operations.
The experimental results present the best performance achieved by ConvNeXtS~\cite{liu2022convnet} trained with backgrounds from MIT indoor scenes dataset~\cite{quattoni2009recognizing} and demonstrate the validity of the obtained dataset on training deep neural networks for deformable wire harness bag segmentation under practical configurations in the actual production.

\subsection{Research regarding wire harness assembly operations}

The identified studies can also be synthesized regarding which operation of wire harness assembly the proposed vision systems contributed to, as shown in Table~\ref{tab:method_diff_oper}.
The majority of previous research discussed the application of computer vision techniques for facilitating the assembly procedure, including four studies on the fixing of clamps~\cite{koo2008development,jiang2010robotized,jiang2011robotized,jiang2015robotized} and seven studies on the mating of connectors~\cite{di2009hybrid,sun2010robotic,di2012vision,tamada2013high,song2017electric,yumbla2020preliminary,zhou2020practical}.
The interpretable classification of wire harness branches explored in \cite{kicki2021tell} and the recognition method for wires proposed in \cite{guo2022visual} can support robotic routing with a better understanding of the topology of wire harnesses.
\cite{koo2008development}, \cite{jiang2010robotized}, and \cite{jiang2011robotized} proposed to install covers on clamps to facilitate the detection and manipulation of clamps on wire harnesses, which, on the other hand, can also assist the robotic routing of wire harnesses by locating the positions of clamps.

\begin{table}[htbp]
    \caption{Contributions of computer vision techniques in previous studies to different assembly operations.}
    \centering
        \begin{tabular}{ll}
            \toprule
            Assembly operation & Articles                                                                                                                                                                                            \\
            \midrule
            Prepare            & \cite{zagar2023copy}                                                                                                                                                                                \\
            Transport          & -                                                                                                                                                                                                   \\
            Untangle           & \cite{zhang2023learning}                                                                                                                                                                            \\
            Route              & \cite{koo2008development,jiang2010robotized,jiang2011robotized,wnuk2021tracking,kicki2021tell,guo2022visual,zurn2023topology}                                                                       \\
            Assemble           & \cite{koo2008development,jiang2010robotized,jiang2011robotized,jiang2015robotized,di2009hybrid,sun2010robotic,di2012vision,tamada2013high,song2017electric,yumbla2020preliminary,zhou2020practical} \\
            \bottomrule
        \end{tabular}
    \label{tab:method_diff_oper}
\end{table}

However, as summarized in Section~\ref{subsec:operation}, there are also other operations within the complete wire harness assembly onto vehicles but gaining less attention, such as preparation, transport, and disentanglement~\cite{zhang2023learning,zagar2023copy}.
Since the wire harness is processed as a pack in preparation and transport, the vision system's task will mainly be identifying the pack of wire harnesses and the location for dropping it.
The untangling of wire harnesses leads to more problems requiring future research.
The robot system needs a dynamic robotic manipulation strategy to react to wire harness deformation, which stresses the significance of real-time wire harness tracking.

\subsection{Vision system evaluation in previous literature}

All the analyzed articles contained experimental studies in laboratory configurations and conducted various evaluations on their proposals~\cite{koo2008development,jiang2010robotized,jiang2011robotized,jiang2015robotized,di2009hybrid,sun2010robotic,di2012vision,tamada2013high,song2017electric,yumbla2020preliminary,zhou2020practical,kicki2021tell,guo2022visual,zhang2023learning,zagar2023copy}.
Specifically, there were various metrics for evaluating the performance of the vision systems in some of the proposals qualitatively and quantitatively, as summarized in Table~\ref{tab:kpi_qual} and Table~\ref{tab:kpi_quan}, respectively.
However, some studies adopted vision systems without reporting the evaluation of their vision systems explicitly~\cite{koo2008development,jiang2010robotized,jiang2011robotized,jiang2015robotized,sun2010robotic,di2012vision,tamada2013high,song2017electric}.

\begin{table}[htbp]
    \caption{Qualitative evaluation on vision systems in previous studies.}
    \centering
        \begin{tabular}{ccl}
            \toprule
            Article                      & Component        & Metrics                                                       \\
            \midrule
            \cite{yumbla2020preliminary} & Connector        & Bounding box, position reference                              \\
            \cite{zhou2020practical}     & Connector        & Bounding box, pose                                            \\
            \cite{kicki2021tell}         & Wire             & CAM~\cite{zhou2016learning}-based class-agnostic saliency map \\
            \cite{guo2022visual}         & Wire             & Recognition result                                            \\
            \cite{zhang2023learning}     & Wire             & Action-grasp pair                                             \\
            \cite{zagar2023copy}         & Wire harness bag & Segmentation mask                                             \\
            \bottomrule
        \end{tabular}
    \label{tab:kpi_qual}
\end{table}

\begin{table*}[htbp]
    \caption{Quantitative evaluation on vision systems in previous studies.}
    \centering
        \begin{tabular}{ccl}
            \toprule
            Article                  & Component        & Metrics                                                                        \\
            \midrule
            \cite{di2009hybrid}      & Connector        & Fault detection rate                                                           \\
            \cite{zhou2020practical} & Connector        & Average Precision (AP); evaluation of repeatability; average and max-min error \\
            \cite{kicki2021tell}     & Wire             & Classification accuracy; confusion matrix                                      \\
            \cite{guo2022visual}     & Wire             & Accuracy; time cost                                                            \\
            \cite{zagar2023copy}     & Wire harness bag & Dice coefficient; Intersection-over-Union (IoU)                                \\
            \bottomrule
        \end{tabular}
    \label{tab:kpi_quan}
\end{table*}

\cite{koo2008development}, \cite{jiang2010robotized}, \cite{jiang2011robotized}, \cite{jiang2015robotized}, \cite{song2017electric}, and \cite{zhang2023learning} qualitatively indicated the technical feasibility of their vision systems for facilitating the robotized manipulation of clamp covers and estimating the relative position of the connector header, respectively, by the execution of robotic manipulation.
\cite{yumbla2020preliminary} demonstrated the qualitative evaluation of their vision system by presenting the detection results with bounding boxes around detected connectors and position references of detected connectors in a simulation environment.
\cite{di2009hybrid} reported the fault detection rate to quantitatively reflect their proposed vision system's performance.
\cite{zhang2023learning} demonstrated the effectiveness of the proposed learning efficient policy for picking entangled wire harnesses with action-grasp pair prediction images.

Besides, \cite{zhou2020practical}, \cite{kicki2021tell}, \cite{guo2022visual}, and \cite{zagar2023copy} reported both qualitative and quantitative experiment results.
\cite{zhou2020practical} demonstrated the performance of detection qualitatively by presenting a bounding box and a pose frame around the detected connector and illustrated the performance and reliability of the proposed algorithm quantitatively with Average Precision (AP), repeatability evaluation, and average and max-min error.
In addition to the qualitative evaluation result with saliency map, \cite{kicki2021tell} also reported classification accuracy and confusion matrix to demonstrate the performance of their proposal quantitatively.
\cite{guo2022visual} evaluated the practicality of their proposal by examining the time cost of each block of their proposed recognition method besides demonstrating the performance of the proposed methods with recognition result images.
Similarly, \cite{zagar2023copy} evaluated the proposed approach regarding the Dice coefficient and IoU and illustrated the performance with segmentation result images.

\section{Discussion}
\label{sec:disc}

In general, previous research has been devoted to enabling robot systems' better autonomy to estimate the state of assembly autonomously for various tasks in wire harness assembly~\cite{salunkhe2023review}.
As a significant aspect in robotics, enabling robotic visual perception has attracted researchers' efforts on robotizing wire harness assembly~\cite{wang2023overview}.
Previous studies explored the application of various computer vision techniques to enable robots to have better visual perception capabilities at different levels of the constituent structure of wire harnesses.
Specifically, previous research mainly focused on the manipulation of different components of wire harnesses~\cite{koo2008development,di2009hybrid,sun2010robotic,jiang2010robotized,jiang2011robotized,di2012vision,tamada2013high,jiang2015robotized,song2017electric,yumbla2020preliminary,zhou2020practical,zagar2023copy}, monitoring sub-processes of the assembly~\cite{di2009hybrid,di2012vision,song2017electric,kicki2021tell,guo2022visual,zhang2023learning}, and fault detection during the assembly~\cite{di2009hybrid}.
These solutions were mainly achieved with vision-based classification and detection on various components of wire harnesses, e.g., clamps~\cite{koo2008development,jiang2010robotized,jiang2011robotized,jiang2015robotized}, connectors~\cite{di2009hybrid,sun2010robotic,di2012vision,tamada2013high,song2017electric,yumbla2020preliminary,zhou2020practical}, wires~\cite{kicki2021tell,guo2022visual,zhang2023learning}, and wire harness bags~\cite{zagar2023copy}.
The proposed vision systems also contributed to different operations regarding current work instructions for wire harness assembly in practical production, including preparation~\cite{zagar2023copy}, disentanglement~\cite{zhang2023learning}, routing~\cite{koo2008development,jiang2010robotized,jiang2011robotized,kicki2021tell,guo2022visual}, and assembly\cite{koo2008development,jiang2010robotized,jiang2011robotized,jiang2015robotized,di2009hybrid,sun2010robotic,di2012vision,tamada2013high,song2017electric,yumbla2020preliminary,zhou2020practical}.
Nevertheless, these existing studies imply challenges and opportunities for future research toward more efficient and practical computer vision-enabled robotized wire harness assembly.

\subsection{Computer vision techniques}

Various industrial manufacturing systems have implemented computer vision techniques to promote informatization, digitization, and intelligence~\cite{zhou2023computer}.
Increasing research efforts in computer vision and robotics have paid attention to the complex DLO manipulation, addressing both theoretical research~\cite{saha2006motion,schulman2013tracking,tang2018framework,wnuk2021tracking,zurn2023topology} and engineering practices~\cite{di2009hybrid,jiang2011robotized,jiang2015robotized,sanchez2018robotic,guo2022visual}.
However, for robotized wire harness assembly, a practical challenge among DLO manipulation problems, it is necessary to recognize wire harnesses successfully within a limited processing time to plan and control robotic manipulation under a constrained production rate~\cite{guo2022visual}.
Besides, estimating the shape of wire harnesses by a pure vision-based algorithm in practical applications in an actual production plant indicates extracting image features from an intricate background~\cite{koo2008development}.
On the other hand, the automotive industry has yet to identify a satisfactory vision-based automation solution that can recognize different components of wire harnesses as robustly as humans.
The inconsistent production configuration in practical manufacturing exacerbates the difficulty of visual recognition.
This has attracted plenty of studies on wire harness component recognition in the past few years~\cite{koo2008development,jiang2010robotized,jiang2011robotized,jiang2015robotized,di2009hybrid,sun2010robotic,di2012vision,tamada2013high,song2017electric,yumbla2020preliminary,zhou2020practical,kicki2021tell,guo2022visual,zhang2023learning,zagar2023copy}.
Nevertheless, further research on object recognition and detection algorithms is needed to achieve a more robust visual machine perception performance for more efficient and practical robotized wire harness assembly.
Specific problems include object detection and pose estimation on wire harness components, topology matching on wire structure, and wire deformation tracking while manipulating wire harnesses in production.

\subsubsection{Utilizing features of wire harnesses}

Novel object recognition and detection algorithms, which harness the intrinsic features of wire harnesses instead of relying on artificial fiducial markers, hold the promise of revolutionizing robotized wire harness assembly, making it more efficient and practical.
Previously, clamp detection was achieved with the assistance of clamp covers with unique designs and markers~\cite{koo2008development,jiang2010robotized,jiang2011robotized,jiang2015robotized}.
Some previous works with connector detection also utilize artificial fiducial markers for object identification and location~\cite{di2009hybrid,sun2010robotic,di2012vision}.

The clamp covers can facilitate recognizing and manipulating clamps with relatively small sizes and complex structures.
The added artificial fiducial markers can simplify feature engineering when designing object recognition algorithms.
However, while effective in recognizing and manipulating clamps, clamp covers present significant challenges in the practical assembly of wire harnesses, particularly in compact installation areas.
Their presence not only occupies valuable space but also complicates the process when they need to be removed, highlighting the need for a more streamlined approach.
It is also not feasible considering an increasing number of wire harnesses assembled in future products, e.g., electric vehicles, as adding artificial fiducial markers would introduce extra complexity, workload, and cost for the assembly operation in actual production~\cite{shi2012mobile}.
In addition, the effectiveness of adding artificial fiducial markers may be impaired due to their various orientations and potential occlusions in actual production environment~\cite{zurn2023topology}.
Therefore, it is desirable to recognize the components of wire harnesses and exploit their intrinsic features without additional attachments.

\subsubsection{Implementing learning-based algorithms}

Earlier research analyzed in this review article mainly achieved 2D image recognition on clamp covers and connectors with traditional rule-based image processing methods~\cite{koo2008development,jiang2010robotized,jiang2011robotized,jiang2015robotized,di2009hybrid,sun2010robotic,di2012vision,tamada2013high,song2017electric,yumbla2020preliminary}.
By contrast, more recent studies on wire recognition took advantage of learning-based algorithms~\cite{zhou2020practical,kicki2021tell,guo2022visual,zhang2023learning,zagar2023copy}, which enabled more robust recognition of objects with a more complex structure, e.g., the flexible and deformable wires and wire harness bags.

The recent renaissance of CNN~\cite{goodfellow2016deep} and the successful development of deep learning in computer vision research~\cite{lecun2015deep} have demonstrated the superior performance of learning-based object recognition than traditional rule-based methods~\cite{zou2023object}.
This booming development of applying deep learning techniques in computer vision research promoted various learning-based designs and solutions for object recognition, e.g., Regions with CNN features (R-CNN)~\cite{girshick2014rich}, Fast R-CNN~\cite{girshick2015fast}, Faster R-CNN~\cite{ren2015faster}, and You Only Look Once (YOLO)~\cite{redmon2016you}.
Applying these learning-based techniques will facilitate better visual machine perception in the future robotic assembly of wire harnesses.

The booming development of AI-supported tools, such as ChatGPT, demonstrates the superiority of large language models (LLM) and generative pre-trained transformers (GPT) in enabling machine intelligence.
The robotics community speedily seized the potential opportunities due to the advent of ChatGPT and initiated the discussion on ``RobotGPT'' to take advantage of LLM and GPT in robotics~\cite{jin2024robotgpt}.
Previously, researchers investigated the strengths of language/text parsing of LLM and proposed various approaches to improving the robotic capability on task planning~\cite{jin2024robotgpt} and human-robot interaction~\cite{xie2023chatgpt}.
Nevertheless, with the advent of large multimodal models (LMM) competent in some visual problems (e.g., GPT-4 with Vision\footnote{\url{https://platform.openai.com/docs/guides/vision}}), the impact of LMM in addressing visual problems in robotic assembly of wire harnesses is worth further investigations~\cite{hou2024more}.

It is also promising to promote the implementation of depth or other 3D cameras to acquire spatial information and process 3D or multi-modality data directly with learning-based methods, considering the increasingly advancing and affordable photography technology and recent development of learning-based object recognition and detection algorithms in the computer vision community~\cite{guo2021deep,zhang2022deep,zou2023object}, 
Therefore, further research is worth conducting on adapting learning-based detection and pose estimation of wire harness components based on multi-modality data.

Nonetheless, the dataset is essential for learning-based object detection~\cite{everingham2009pascal,lin2014microsoft,russakovsky2015imagenet} and scalable deep learning-based computer vision applications in manufacturing~\cite{nguyen2022enabling,zhou2023computer}.
Additionally, benchmarks are critical to assess and compare the performance across different computer vision and robotic systems~\cite{kimble2022performance}.
Though increasingly popular within the robotics community, benchmarks for various tasks have yet to gain sufficient research for robotized wire harness assembly~\cite{wang2023deep}.
Thus, in addition to learning networks, more studies are needed on the datasets, benchmarks, and metrics to promote the performance of learning-based computer vision techniques implemented for future robotized wire harness assembly and evaluate them consistently and rigorously.

\subsubsection{Learning from multi-modality data}

Previous research efforts summarized in Table~\ref{tab:clamp}, Table~\ref{tab:connector}, and Table~\ref{tab:wire} indicate the dominant adaptation of 2D vision in previous research on facilitating robotized wire harness assembly with computer vision techniques.
For example, for connectors, recognition was mainly used to provide geometric information to the control system so that the robot could flexibly perceive and manipulate the connectors~\cite{di2009hybrid,sun2010robotic,di2012vision,tamada2013high,song2017electric,yumbla2020preliminary,zhou2020practical}.
Previous research mainly adapted 2D image recognition to acquire the positions or orientations of connectors~\cite{di2009hybrid,sun2010robotic,di2012vision,tamada2013high,song2017electric,zhou2020practical}.
For example, \cite{tamada2013high}, using a 2D camera, assumed that all connectors were placed on a flat workbench to reduce the degrees of freedom of connectors.
However, considering the actual manufacturing scenario, wire harnesses are not fixed on a flat workbench before or during assembly, making the connectors distributed randomly in the 3D space.
Hence, the degrees of freedom of connectors cannot be reduced directly, which results in limited practicality of the vision system proposed by \cite{tamada2013high} in actual production.
\cite{sun2010robotic} implemented two 2D cameras to ensure the correct positioning of connectors in 3D space, which increases the complexity of vision system control.
Therefore, it is desired to collect 3D information directly, e.g., depth information or point clouds of objects of interest, to better perceive wire harness components.

The recent acceleration in the development of depth cameras and 3D scanners makes it more achievable to acquire and process depth or 3D visual information in addition to 2D data, e.g., RGB or grayscale images.
However, though an RGB-D camera was adopted, \cite{yumbla2020preliminary} acquired the positions of connectors based on 2D image processing and mapped the identified positions to the registered depth image to further obtain the 3D measurement of connectors instead of processing depth or other 3D information independently or together with 2D color information.
The more recent studies on recognizing the wire part explored implementing RGB-D camera or 3D scanner and learning from depth images or 3D data and have demonstrated the performance of 3D vision applications on facilitating a better perception of the structure of wire harnesses~\cite{kicki2021tell,guo2022visual,zhang2023learning}.
On the one hand, these results enable robotized preparation, transport, disentangling, and routing of wire harnesses in the future assembly station.
On the other hand, they provide references to address computer vision applications on the other components of wire harnesses based on depth images, point clouds, or other 3D data.

\subsubsection{Adapting studies for wire harness manufacturing}

It is noteworthy that there are also studies referring ``wire harness assembly'' to the operations in the manufacturing process of wire harnesses~\cite{trommnau2019overview,nguyen2021manufacturing,navas2022wire,makris2023automated}.
Manufacturing wire harnesses is out of the scope of this systematic review.
Nevertheless, some sub-problems are shared among computer vision applications in the assembly of wire harnesses and the assembly of wire harnesses onto other products.
For example, in both scenarios, detecting components of wire harnesses and state estimation during wire manipulation is required to provide robots with the necessary information before conducting the following robotic grasping and manipulation.

Notably, previous research has explored vision-based solutions for robotic manufacturing of wire harnesses or wire harness assembly processes.
These innovative designs on vision systems could serve as a valuable reference for the robotized wire harness assembly discussed in this article, hinting at promising avenues for future exploration.
Therefore, it is imperative that future research focuses on validating the effectiveness of vision systems proposed for facilitating the robotic manufacturing of wire harnesses in the robotic installation of wire harnesses onto other products.

\subsection{Practical concerns regarding actual production}

The previous research efforts in vision-based robotized wire harness assembly remained experimental studies in controlled laboratory settings.
However, vision systems must address various practical conditions to be implemented in industrial production, which poses practical concerns for developing vision-based solutions in future research.

\subsubsection{Evaluating vision systems regarding production requirements}
\label{subsubsec:evaluating}

The analyzed articles have demonstrated various vision-based solutions for accomplishing different tasks of robotized wire harness assembly under laboratory settings~\cite{koo2008development,jiang2010robotized,jiang2011robotized,jiang2015robotized,di2009hybrid,sun2010robotic,di2012vision,tamada2013high,song2017electric,yumbla2020preliminary,zhou2020practical,kicki2021tell,guo2022visual,zhang2023learning,zagar2023copy}.
However, assessing the proposed vision systems in actual production remains obscure but necessary before they are integrated.
Table~\ref{tab:kpi_qual} and Table~\ref{tab:kpi_quan} present the qualitative and quantitative evaluations conducted on vision systems in different studies.
Nevertheless, only a few studies considered the practical issues, e.g., repeatability of the proposal~\cite{zhou2020practical} and time cost on the vision system~\cite{guo2022visual}.

The evaluation of the proposed vision systems is critical to selecting vision systems, both hardware and software, in practical industrial applications~\cite{perez2016robot}.
Quantitatively evaluating proposed vision systems regarding specific production requirements can provide practitioners with valuable suggestions on selecting the appropriate vision-based solutions according to their respective needs.
As shown in Table~\ref{tab:clamp}, Table~\ref{tab:connector}, and Table~\ref{tab:wire}, various types of vision systems were adopted in previous research in robotized wire harness assembly, including stereo vision, monocular vision, and the combination of them, for different purpose, e.g., obtaining positions and orientations of wire harness components or compensating occlusion from specific views.
Different types of cameras were also discussed on the level of devices, e.g., RGB cameras, depth cameras, and 3D scanners.
However, as shown in Table~\ref{tab:kpi_quan}, a few studies conducted quantitative evaluations on the performance of proposed vision systems.
This indicates a need for more consideration of the practicality of vision systems in actual industrial applications.
In conclusion, the research underscores the need for comprehensive evaluations of different vision systems in the context of practical production requirements.
This type of evaluation is essential to provide practitioners with more robust and practical advice on selecting vision systems for their specific industrial applications.
Therefore, further research in this direction is highly recommended.

\subsubsection{Considering actual production environment}

In a practical manufacturing environment, it is inevitable for the vision system to cope with inconsistent conditions, e.g., background~\cite{koo2008development} and illumination conditions~\cite{jiang2011robotized}.
\cite{koo2008development} revealed the complex background as a non-negligible hindrance for a pure vision-based wire shape estimation in an actual plant.
\cite{jiang2011robotized} indicated the lack of vision and laser processing robustness as one of the significant reasons behind experiment failures, e.g., the variation of illumination conditions.
Thus, it is critical to evaluate the practicality, robustness, and reliability of vision systems regarding different perspectives, e.g., recognition accuracy, processing speed, and physical facilities in the assembly station, under actual production configurations with various background and illumination conditions.
\cite{zagar2023copy} initiated the consideration of practical operation background from actual production in evaluating the proposed vision-based solution, and more attention from the research community should be paid to this perspective.

Another practical issue that has not been discussed previously is the fact that some wire harnesses are installed onto the final product on a moving production line~\cite{shi2012mobile}, for which \cite{shi2012mobile} proposed a method for a mobile robot manipulator assembling wire harnesses to track a moving vehicle with visual servoing.
Nevertheless, the vision system's processing time and feature engineering process are still challenging considering actual production requirements~\cite{shi2012mobile}.
Hence, more studies are required to address the robotized wire harness assembly on a moving production line.

\subsubsection{Fulfilling economic requirements}

From the business perspective, there is also a demand for the processing time of robotic assembly to fulfill the productivity in practical production, which requires the vision system to perceive the state of the manipulating object fast enough to allow for a following robot action~\cite{bodenhagen2014adaptable}.
\cite{jiang2010robotized} demonstrated the technical feasibility of robotized wire harness assembly, but the average speed and reliability were still far from the requirement of practical application.
\cite{guo2022visual} also identified the necessity of promoting time efficiency.
Therefore, evaluating the vision system from the perspective of processing time is desired and necessary, which, however, was involved little in previous research~\cite{guo2022visual}.
Regarding sustainability from the economic and social perspectives, it is also critical for practical application to consider the pay-off of automating the overall or part of current manual wire harness assembly by implementing vision systems~\cite{aguirre1994economic,aguirre1997robotic,kimble2022performance}.

\subsection{Human-robot collaboration}

The deformability of wire harnesses has been identified as a significant problem for automating the assembly of wire harnesses regarding the perception and manipulation~\cite{koo2008development}.
Automation has been widely adopted in manufacturing since the third industrial revolution~\cite{leng2022industry}.
Nevertheless, the lack of flexibility and cognitive ability in robots motivates the study of human-robot collaboration (HRC)~\cite{wang2020overview}, where the system can benefit from the synergy of humans' strength in perception and flexibility and robots' superiority in payload, accuracy, and repeatability.
Recently, Industry~5.0 has also been initiated with a focus on sustainability, human-centricity, and resilience~\cite{euro2020enabling,euro2021industry,euro2022industry}.
This further promotes the discussions on implementing human-robot collaboration in industrial applications towards human-centric automation~\cite{wang2020overview}.

There have been studies on human-robot collaboration driven by computer vision techniques in industrial applications~\cite{mohammed2017active} and particularly in handling deformable objects~\cite{kruse2015collaborative}.
\cite{heisler2021optimization} also proposed to optimize wire harness assembly based on human-robot collaboration.
However, previous research on computer vision-based robotized wire harness assembly focused on fully robotic assembly~\cite{koo2008development,jiang2010robotized,jiang2011robotized,jiang2015robotized,di2009hybrid,sun2010robotic,di2012vision,tamada2013high,song2017electric,yumbla2020preliminary,zhou2020practical,kicki2021tell,guo2022visual,zhang2023learning}.
More research is needed to investigate human-centered automation and human-robot collaboration for the robotic assembly of wire harnesses.

To introduce human-robot collaboration to robotized wire harness assembly, several challenges demand further research, e.g., task allocation and addressing a safe human-robot interaction~\cite{salunkhe2023specifying}.
This makes it crucial to design the workspace to maximize the efficiency of human-robot collaboration while minimizing the risk of accidents.
Conventionally, industrial robots are surrounded by physical fences or laser curtains in practical automated production lines to ensure the safety of the operation~\cite{iso10218part1,iso10218part2}.
However, in human-robot collaboration, a human operator and a robot work at a closer distance than the one for current industrial robots, which significantly promotes the priority of ensuring the safety of human operators~\cite{iso15066,gerbers2018safe,kim2021considerations}.
Collaborative robots (or cobots) have been identified as the key enabling technology for better performing various automation tasks by combining the skills of humans and robots while maintaining safety and efficiency~\cite{hentout2019human,proia2022control}.
Therefore, applying cobots can be a promising solution for automating wire harness assembly.
Various computer vision techniques for facilitating an efficient and safe human-robot collaboration have also been discussed previously, such as collision detection, environment perception, human action or gesture recognition, and proactive human-robot collaboration~\cite{li2021towards,li2021toward,fan2022vision,li2023proactive}.
These studies pave the way for robotized wire harness assembly with computer vision-driven human-robot collaboration.
Recent advance on ``RobotGPT'' also suggests novel directions for developing efficient, effective, and safe human-robot collaboration with LLM or LMM for robotized wire harness assembly~\cite{xie2023chatgpt,ye2023improved}.

\subsection{Product design of wire harnesses}

Previously, \cite{koo2008development} identified that, though the clamps could be regarded as rigid objects, their relatively small sizes and complex shapes made clamps challenging to visually recognize and grasp directly by a robot gripper.
Therefore, clamp covers with markers were installed to facilitate the detection of clamps~\cite{koo2008development,jiang2010robotized,jiang2011robotized,jiang2015robotized}.
However, affixing clamp covers will be impractical in future production, where an increasing number of wire harnesses will be installed in tight areas.
Besides, various intrinsic properties of current wire harnesses, e.g., the same color of clamps and taped wires and the small radial scale and complex structure of wire harnesses with irregular curves and crossings, also make it considerably challenging to recognize wire harnesses from the complex background and assemble automatically~\cite{koo2008development,guo2022visual}.
Hence, inspired by the ``Design for X (DfX)'' philosophy~\cite{boothroyd2005assembly,chiu2010evolution}, novel designs of wire harness components are desired to facilitate visual detection without any additional parts attached to wire harnesses.

\subsection{Industrialization barriers}
\label{subsec:barriers}

Consciously, the research in robotized wire harness assembly is task-specific and application-oriented.
Thus, the research envisions the industrialization of proposed technical solutions.
However, implementing automation solutions in wire harness assembly remains scarce in production.
This situation could be caused by 1) the lack of interest in industry and/or 2) the unsatisfied research results and/or 3) barriers to transferring research results to industrial practice.
The industry has exhibited remarkable interest in robotizing the assembly of wire harnesses~\cite{salunkhe2023review,wang2023overview}.
However, as discussed in previous sections of this article, the research is insufficient, and the proposed technical solutions are not powerful enough, so this review advocates future research directions.
Nevertheless, several aspects related to industrial implementation should also be addressed besides improving robotic systems to facilitate industrialization, e.g., cost, workforce competence, trustworthiness, and standardization~\cite{zhu2022challenges,makris2023automated}.

Cost is fundamental in researching, developing, commercializing, and operating technologies.
To reach a satisfactory level, vast investment in funding and human efforts is required to study and develop vision-based solutions.
Meanwhile, it is critical to balance the performance and cost of technical solutions in practical operations.
This requires controlling the price of hardware and software and keeping operations' complexity to manageable levels.
The engineering cost of commercializing the technology is also significant.
Thus, substantial efforts are required to investigate how to do so efficiently.

Considering workforce competence, training operators on new skills is inevitable when introducing new technologies~\cite{singer2023managing}.
This poses requirements on motivating and training operators fast and with high quality, where research in related challenges and treatments are in need~\cite{braun2022motivational}.
As a prerequisite, trust in new technologies is crucial to accepting technologies, especially for AI-empowered automation solutions.
This socio-ethical issue drives academia to promote the explainability and trustworthiness of AI to address the lack of trust in AI-based solutions in the industry.
The growing implementation of AI-based technologies also spawns concerns about cybersecurity and data management, which requires research in relevant fields~\cite{lee2018industrial,tuptuk2018security}.

Additionally, standardization is elemental to new technologies.
Although academia has initiated related research, there is no widely applicable standard or benchmark for vision-based robotic manipulation of DLOs in practical industrial production~\cite{kimble2022performance}.
This situation calls for further research on methods for benchmarking the performance and evaluation of vision-based solutions in academia.
In the meantime, researchers, practitioners, and policymakers should also discuss legislating the certification of existing research results with standards and determining requirements to guide future research and engineering.

\section{Conclusion}
\label{sec:con}

Robotized wire harness assembly is desired in the automotive industry, considering its strength in promoting assembly quality, productivity, safety, and ergonomics.
However, it is challenging due to the flexibility and deformability of wire harnesses, the small sizes and complex shapes of wire harness components, and the demanding production requirements.

Through a systematic literature review, this article revealed that previous research explored various computer vision techniques for facilitating robotized wire harness assembly with better robotic visual perception of the wire harness components to be manipulated and the manipulation operations.
Different vision-based solutions were proposed for various sub-tasks of robotized wire harness assembly, including manipulating clamps and connectors, identifying wire harness bags, monitoring the mating process of connectors, fault detection during assembly, and bin-picking problem by detecting different wire harness components, including clamps, connectors, wires, and wire harness bags.


Based on past research, this article identified two major challenges for computer vision applications in the robotic assembly of wire harnesses:

\begin{enumerate}
    \item The robustness of vision systems in actual production has not achieved the compatible level as humans, especially considering the demanding production rate and intricate production environments;
    \item Intrinsic physical properties of different wire harness components were identified as hindrances to robotic visual perception.
\end{enumerate}

Furthermore, this article proposed prospective research directions toward more efficient and practical computer vision applications in robotized wire harness assembly:

\begin{enumerate}
    \item Developing and implementing learning-based computer vision techniques to exploit intrinsic features and multi-modality data of wire harnesses
    \item Investigating the adaptation of computer vision techniques proposed for the robotized assembly process of manufacturing wire harnesses
    \item Evaluating the practicality, robustness, reliability, and sustainability of the proposed vision systems regarding practical manufacturing scenarios
    \item Inquiring novel vision system designs considering human-robot collaborations and different assembly operations
    \item Exploring new wire harness designs for facilitating more efficient visual recognition
\end{enumerate}

This article also advocates addressing barriers to the industrialization of technologies, which demands investigation and discussion among researchers and practitioners from various backgrounds.

\section*{Acknowledgements}

This work was supported by the Swedish innovation agency, Vinnova, and the strategic innovation program, Produktion2030, under grant number 2022-01279.
The work was carried out within Chalmers Production Area of Advance.
The support is gratefully acknowledged.



\bibliographystyle{elsarticle-num}
\bibliography{mybib}

\end{document}